\documentclass[11pt]{article}

\usepackage{array}
\newcolumntype{P}[1]{>{\centering\arraybackslash}p{#1}}

\usepackage{booktabs} 

\usepackage{hyperref}

\usepackage{tikz}
\usepackage{verbatim}

\usepackage{amssymb}
\usepackage{lineno}
\usepackage{mathtools,bbm}
\usepackage{graphicx,url}
\usepackage{graphics}

\DeclareMathOperator*{\E}{\mathbb{E}}
\DeclareMathOperator*{\R}{\mathcal{R}}

\DeclareMathOperator*{\F}{\mathcal{F}}
\DeclareMathOperator*{\X}{\mathcal{X}}

\DeclareMathOperator*{\argmin}{argmin}

\usepackage{setspace,graphicx,epstopdf,amsmath,amsfonts,amssymb,amsthm,versionPO}
\usepackage{marginnote,datetime,enumitem,rotating,fancyvrb,scalerel}
\usepackage{hyperref}

\excludeversion{notes}		
\includeversion{links}          

\iflinks{}{\hypersetup{draft=true}}

\ifnotes{%
\usepackage[margin=1in,paperwidth=10in,right=2.5in]{geometry}%
\usepackage[textwidth=1.4in,shadow,colorinlistoftodos]{todonotes}%
}{%
\usepackage[margin=1in]{geometry}%
\usepackage[disable]{todonotes}%
}

\makeatletter\let\chapter\@undefined\makeatother 

\usepackage{indentfirst} 
\usepackage{jfe}          

\newtheorem{proposition}{Proposition}

\newtheorem{definition}{Definition}

\usepackage[
  backend=biber,
  style=numeric, 
]{biblatex}
\usepackage{float}
\usepackage[caption = false]{subfig}

\begin{document}

\setlist{noitemsep}  

\title{\large\textbf{Deep Neural Networks for Choice Analysis: \\ Architecture Design with Alternative-Specific Utility Functions}}

\author{Shenhao Wang \\
  Baichuan Mo \\
  Jinhua Zhao \\
  \\
  Massachusetts Institute of Technology\\}

\date{March 2020\footnote{Please cite: Wang, Mo, and Zhao (2020). "Deep neural networks for choice analysis: Architecture design with alternative-specific utility functions." Transportation research part C: emerging technologies 112: 234-251.}}              

\renewcommand{\thefootnote}{\fnsymbol{footnote}}

\singlespacing

\maketitle

\vspace{-.2in}
\begin{abstract}
\noindent
Whereas deep neural network (DNN) is increasingly applied to choice analysis, it is challenging to reconcile domain-specific behavioral knowledge with generic-purpose DNN, to improve DNN’s interpretability and predictive power, and to identify effective regularization methods for specific tasks. To address these challenges, this study demonstrates the use of behavioral knowledge for designing a particular DNN architecture with alternative-specific utility functions (ASU-DNN) and thereby improving both the predictive power and interpretability. Unlike a fully connected DNN (F-DNN), which computes the utility value of an alternative $k$ by using the attributes of \textit{all} the alternatives, ASU-DNN computes it by using only $k$'s \textit{own} attributes. Theoretically, ASU-DNN can substantially reduce the estimation error of F-DNN because of its lighter architecture and sparser connectivity, although the constraint of alternative-specific utility can cause ASU-DNN to exhibit a larger approximation error. Empirically, ASU-DNN has 2-3\% higher prediction accuracy than F-DNN over the whole hyperparameter space in a private dataset collected in Singapore and a public dataset available in the R mlogit package. The alternative-specific connectivity is associated with the independence of irrelevant alternative (IIA) constraint, which as a domain-knowledge-based regularization method is more effective than the most popular generic-purpose explicit and implicit regularization methods and architectural hyperparameters. ASU-DNN provides a more regular substitution pattern of travel mode choices than F-DNN does, rendering ASU-DNN more interpretable. The comparison between ASU-DNN and F-DNN also aids in testing behavioral knowledge. Our results reveal that individuals are more likely to compute utility by using an alternative’s own attributes, supporting the long-standing practice in choice modeling. Overall, this study demonstrates that behavioral knowledge can guide the architecture design of DNN, function as an effective domain-knowledge-based regularization method, and improve both the interpretability and predictive power of DNN in choice analysis. Future studies can explore the generalizability of ASU-DNN and other possibilities of using utility theory to design DNN architectures.
\end{abstract}
\medskip


\thispagestyle{empty}

\clearpage

\onehalfspacing
\setcounter{footnote}{0}
\renewcommand{\thefootnote}{\arabic{footnote}}
\setcounter{page}{1}

\section{Introduction}
\label{introduction}
\noindent
Choice analysis is an important research area across economics, transportation, and marketing \cite{McFadden1974,Ben_Akiva1985,Guadagni1983}. Whereas discrete choice models were traditionally used to analyze this question, recently researchers have become increasingly interested in applying machine learning (ML) methods such as deep neural network (DNN) to analyze individual choices \cite{Karlaftis2011,Paredes2017,WangShenhao2018_ml2}. While DNN has demonstrated its extraordinary predictive power in the tasks such as image recognition and natural language processing, its application to demand analysis is still hindered by at least three problems. First, as DNN gradually permeates into many domains, it is unclear how generic-purpose DNN classifiers can be reconciled with domain-specific knowledge \cite{LeCun2015, Qianli2018}. Whereas the ML community generally admires the effectiveness of automatic feature learning in DNN \cite{LeCun2015}, heated debates continue with regard to the extent and manner in which domain knowledge can be used to improve ML models and solve domain-specific problems more efficiently \cite{Qianli2018}. Second, because DNN is a significantly more complicated generic-purpose model, its interpretability is generally considered to be low \cite{Lipton2016,Koh2017}. Even though it is relatively straightforward to apply DNN to forecast demand, researchers have obtained limited policy and behavioral insights from DNN until now. Third, even the prediction itself can be challenging because of the high dimensionality and data overfitting of DNN. Effective regularization methods and DNN architectures are important to improve the out-of-sample performance. Whereas many recent progresses were achieved by creating novel DNN architectures, the procedure of designing deep architecture is still largely ad hoc without systematic guidance \cite{ZhangChiyuan2016, Mhaskar2016}. These three challenges, including the tension between domain-specific and generic-purpose knowledge, lack of interpretability, and challenge of identifying effective regularization and architecture, are theoretically important and empirically critical for applying DNN to any specific domains. 

\textit{To address these problems, this study demonstrates the use of behavioral knowledge for designing a novel DNN architecture with alternative-specific utility functions (ASU-DNN), thereby improving both the predictive power and interpretability of DNN in choice analysis.} We first elaborate on the implicit interpretation of random utility maximization (RUM) in DNN, framing the question of DNN architecture design as one of utility specification. This insight results in the design of the new ASU-DNN architecture, in which the utility of an alternative depends only on its own attributes, as opposed to a fully connected DNN (F-DNN) in which the utility of each alternative is the function of all the alternative-specific variables. Using statistical learning theory, we demonstrate that this ASU-DNN architecture can reduce the estimation error of F-DNN thanks to its much sparser connectivity and fewer parameters, although the approximation error of ASU-DNN could be higher. We then apply ASU-DNN, F-DNN, multinomial logit (MNL), nested logit (NL), and nine benchmark ML classifiers to predict travel mode choice by using two datasets, referred to as SGP and TRAIN in this study. The SGP dataset was collected in Singapore in 2017, and the TRAIN dataset was from the mlogit package in R. Our results demonstrate that ASU-DNN exhibits consistently higher prediction accuracy than F-DNN and the other eleven classifiers in predicting travel mode choice over the whole hyperparameter space. The alternative-specific connectivity design in ASU-DNN leads to an IIA-constraint substitution pattern across the alternatives, which can be considered as a domain-knowledge-based regularization, in contrast to the generic-purpose regularization methods such as explicit and implicit regularizations and other architectural hyperparameters. Our results show that the domain-knowledge-based regularization is more effective than the generic-purpose regularization in improving the prediction performance. Finally, we interpret the substitution pattern across travel mode alternatives in ASU-DNN by using sensitivity analysis and demonstrate that ASU-DNN reveals more reasonable behavioral patterns than F-DNN owing to its more regular and intuitive choice probability functions. Overall, the behavioral knowledge of alternative-specific utility function can be used to partially address all three challenges of DNN applications by integrating generic-purpose DNN and domain-specific behavioral knowledge, improving the predictive power and interpretability of ``black box'' DNN, and functioning as an effective domain-knowledge-based regularization.

Broadly speaking, this study points to a new research direction of injecting behavioral knowledge into DNN and enhancing DNN architectures specifically for choice analysis. We aim to advance domain-specific behavioral knowledge using DNN, as opposed to simply applying DNNs for prediction adopted by most recent studies in the transportation domain. This research direction is feasible because the behavioral knowledge used in the classic choice models has a counterpart in the DNN architecture. Specifically, the substitution pattern between alternatives can be controlled by the connectivity of the DNN architecture, and vice versa. From an ML perspective, behavioral knowledge can function as domain-knowledge-based regularization, which better fits domain-specific tasks than generic-purpose regularizations. The alternative-specific utility is only one small piece in the rich set of behavioral insights accumulated over decades of transportation scholarship, and future studies can explore and create more noteworthy DNN architectures for choice analysis based on this behavioral perspective. To facilitate future research, we uploaded this work to a Github repository: \url{https://github.com/cjsyzwsh/ASU-DNN.git}.

The paper is organized as follows: The next section reviews studies on DNN’s applications, interpretability, and regularization methods. Section 3 examines three theoretical aspects of DNN: the relationship between RUM and DNN, architecture design of ASU-DNN, and estimation and approximation error tradeoff between ASU-DNN and F-DNN. Section 4 presents the experiments, and discusses the prediction accuracy, effectiveness of domain-knowledge-based regularization, and interpretability of ASU-DNN. Section 5 concludes.

\section{Literature Review}
\noindent
Individual decision-making has been an important topic in many domains, including marketing \cite{Guadagni1983}, economics \cite{McFadden1974}, transportation \cite{Ben_Akiva1985,Train2009}, biology \cite{Sham1995}, and public policy \cite{Borsch1988}. In recent years as ML models permeated into these domains, researchers started to use various classifiers to analyze how individuals take decisions \cite{Paredes2017, Karlaftis2011}. In the transportation domain, Karlaftis and Vlahogianni (2011) \cite{Karlaftis2011} summarized the transportation fields in which DNN models are used, including (1) traffic operations (such as traffic forecasting and traffic pattern analysis); (2) infrastructure management and maintenance (such as pavement crack modeling and intrusion detection); (3) transportation planning (such as in travel mode choice and route choice modeling); (4) environment and transport (such as air pollution prediction); (5) safety and human behavior (such as accident analysis); and (6) air, transit, rail, and freight operations. Recently, many studies applied SVM, decision tree (DT), RF, and DNN to predict travel behavior, automobile ownership, traffic accidents, traffic flow, or even travelers' decision rules \cite{Pulugurta2013,Omrani2015,Sekhar2016,Paredes2017,Cantarella2005,Polson2017,LiuLijuan2017,ZhangZhenhua2018,Cranenburgh2019}. However, nearly all of these studies apply certain generic-purpose ML models to solve domain-specific transportation problems, but none of them explored how domain-specific knowledge could be used to improve generic-purpose ML models for specific tasks. 

The balance between generic-purpose DNN classifiers and domain-specific knowledge is a general challenge to the application of DNN to any specific domain. On the one hand, DNN is effective owing to its generic-purpose and automatic feature learning capacity \cite{LeCun2015,Bengio2013}. For example, the hyperparameters and architecture in feedforward neural network such as ReLU activation functions can be widely used regardless of the differences between natural language processing (NLP), image recognition, and travel behavioral analysis \cite{Krizhevsky2012, Szegedy2015}. On the other hand, a few studies indicate that handcrafted features could still aid in constructing DNN models \cite{Qianli2018}. In fact, certain domain-specific knowledge is generally involved in DNN modeling. For example, the use of max pooling layer or data augmentation in CNN relies on our domain-specific understanding of images, such as their invariance properties \cite{Goodfellow2016}.

Another challenge to DNN application is DNN's lack of interpretability, which is caused by its complex model assumptions \cite{Lipton2016,Boshi_Velez2017}. The interpretability of DNN is particularly important for reasons such as safety, transparency, trust, and construction of new knowledge \cite{Freitas2014, Brauneis2017}. The majority of the ML studies applied to the transportation field focus exclusively on prediction, which is valid because ML models were initially designed for prediction \cite{Nijkamp1996,Rao1998,XieChi2003,Omrani2015,Hagenauer2017}. Prediction-driven ML models differ significantly from the classical choice models, which are both predictive and interpretable \cite{McFadden1974}. However, to describe DNN as totally a ``black-box'' may be biased because many recent studies have demonstrated various methods of interpreting DNN. These methods could be categorized broadly into two: ex-ante interpretation \cite{Ribeiro2016} (which improves interpretability before model building) and post-hoc interpretation (which focuses on extracting information after model training) \cite{Boshi_Velez2017}. For example, CNN can be interpreted in a post-hoc manner by visualizing the semantic contents in image recognition tasks \cite{ZhouBolei2014}. In choice analysis, it appears feasible to post-hoc interpret DNN and derive the economic information from DNNs \cite{WangShenhao2018_ml2,Rao1998,Bentz2000}. Some other studies used the computational graphs to represent the travel demand structures \cite{WuXin2018,SunJianping2019}. However, these studies that use the visualization of computational graphs did not examine the connection between the utility theory that the choice modeling relies on and the compositional structure of the hidden layers that is the hallmark of DNNs \cite{Poggio2017}, failing to take advantage of either the function approximation capacity of DNNs or the rigorous behavioral insights captured in utility theories.

Even only for prediction, it is significantly challenging to design effective regularization methods and DNN architectures. The regularization methods in DNN consist of explicit and implicit ones, and recent studies reveal that explicit regularizations such as $l_1$ and $l_2$ penalties may not effectively aid in the generalization of DNN \cite{ZhangChiyuan2016}. New DNN architectures could also aid in improving DNN performance. Recent studies either manually design new architectures (such as AlexNet \cite{Krizhevsky2012}, GoogLeNet \cite{Szegedy2015}, and ResNet \cite{HeKaiming2016}) or automatically search for novel architectural design by using Gaussian process, reinforcement learning, or other sequential modeling techniques \cite{Snoek2015,Joz2015,Zoph2016,Falkner2018}. However, most architecture designs are ad hoc explorations without systematic guidance, and the final DNN architecture identified through automatic searching is not interpretable.

\section{Theory}
\subsection{Random Utility Maximization and Deep Neural Network}
\noindent
There are two types of inputs in choice modeling: alternative-specific variables $x_{ik}$ and individual-specific variables $z_i$. Using travel mode choice as an example: $x_{ik}$ could be the price of different travel modes, and $z_i$ represents individual characteristics, such as income and education. $i \in \{ 1, 2, ... N \}$ is the individual index, and $k \in \{ 1, 2, ... K\}$ is the alternative index. Let $B = \{ 1, 2, ... K\}$ and $\tilde x_i =[x^T_{i1}, ..., x^T_{iK}]^T$. The output of choice modeling is individual $i$'s choice, denoted as $y_i = [y_{i1}, y_{i2}, ... y_{iK}]$. Each $y_{ik} \in \{0,1\}$ and $\underset{k}{\sum} y_{ik} = 1$. RUM assumes that the utility of each alternative is the sum of the deterministic utility $V_{ik}$ and random utility $\epsilon_{ik}$:

\begin{equation}
\setlength{\jot}{2pt} \label{eq:util}
  \begin{aligned}
  U_{ik} = V_{ik}(z_i, \tilde x_i) + \epsilon_{ik}
  \end{aligned}
\end{equation}

\noindent
Individuals tend to select the maximum utility out of $K$ alternatives with probabilities. The probability that individual $i$ selects alternative $k$ is

\begin{equation}
\setlength{\jot}{2pt} \label{eq:prob}
  \begin{aligned}
  P_{ik} &= Prob(V_{ik} + \epsilon_{ik} > V_{ij} + \epsilon_{ij}, \forall j \in B, \ j \neq k)
  \end{aligned}
\end{equation}

\noindent
Assuming that $\epsilon_{ik}$ is independent and identically distributed across individuals and alternatives and that the cumulative distribution function of $\epsilon_{ik}$ is $F(\epsilon_{ik})$, the choice probability

\begin{equation}
\setlength{\jot}{2pt} \label{eq:prob_1}
  \begin{aligned}
  P_{ik} &= \int \underset{j \neq k}{\prod} F_{\epsilon_{ij}}(V_{ik} - V_{ij} + \epsilon_{ik}) d F(\epsilon_{ik})
  \end{aligned}
\end{equation}

The following two propositions demonstrate how DNN and RUM are related. The proof of the two propositions is available in Appendix I.

\begin{proposition} \label{prop:1}
Suppose $\epsilon_{ik}$ follows the Gumbel distribution, with probability density function equals to $f(\epsilon_{ik}) = e^{-\epsilon_{ik}} e^{-e^{-\epsilon_{ik}}}$ and  cumulative distribution function equals to $F(\epsilon_{ik}) = e^{-e^{-\epsilon_{ik}}}$. Then, the choice probability $P_{ik}$ takes the form of the Softmax activation function $P_{ik} = \frac{e^{V_{ik}}}{\underset{j}{\sum} e^{V_{ij}}}$.
\end{proposition}

\noindent
The proof is available in many choice modeling textbooks \cite{Train2009,Ben_Akiva1985}.

\begin{proposition} \label{prop:2}
Suppose that Equation \ref{eq:prob_1} holds and that choice probability $P_{ik}$ takes the form of Softmax function as in Equation \ref{eq:choice_prob_deriv}. If $\epsilon_{ik}$ is a distribution with the transition complete property, $\epsilon_{ik}$ follows the Gumbel distribution, with $F(\epsilon_{ik}) = e^{-\alpha e^{-{\epsilon}_{ik}}}$.
\end{proposition}

\noindent
The proof is available in lemma 2 of McFadden (1974) \cite{McFadden1974}. 

Propositions 1 and 2 illustrate the close relationship between RUM and DNN. When F-DNN is applied to the inputs $\tilde x_i$ and $z_i$, the implicit assumption is of RUM with a random utility term following the Gumbel distribution. The inputs into the Softmax function in the DNN could be interpreted as utilities of alternatives. The Softmax function itself could be considered as a soft method of comparing utility scores. The DNN transformation prior to the Softmax function could be considered as the process of specifying utilities.

\begin{figure}[ht]
\centering
\includegraphics[width=0.3\linewidth]{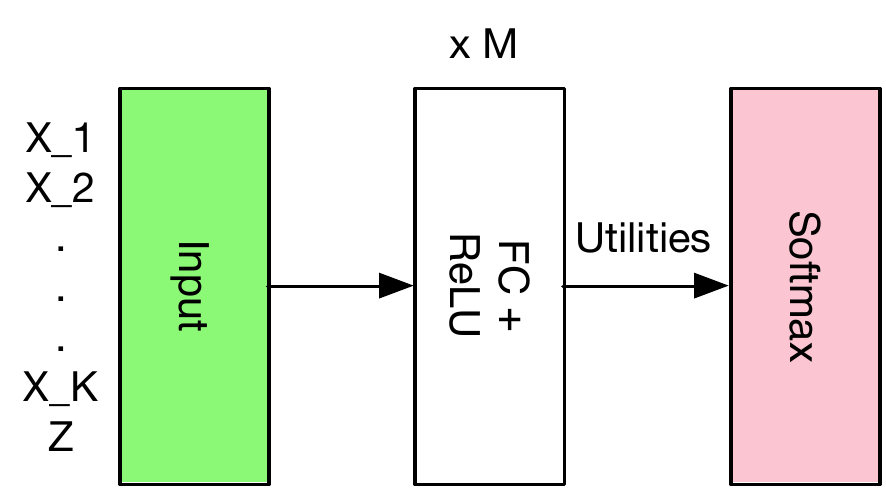}
\caption{Fully Connected Feedforward DNN (F-DNN); it is a standard feedforward DNN. The inputs incorporate both alternative-specific and individual-specific variables. The inputs into the Softmax activation function can be interpreted as utilities.}
\label{fig:arch1}
\end{figure}

Formally, $V_{ik}$ in F-DNN follows:

\begin{equation}
\setlength{\jot}{2pt} \label{eq:util_specification}
  \begin{aligned}
  V_{ik} &= V(z_i, \tilde x_i) = w_k^T \Phi(z_i, \tilde x_i) = w_k^T (g_{m} ... \circ g_2 \circ g_1)(z_i, \tilde x_i) \\
  \end{aligned}
\end{equation}

\noindent
$m$ is the number of layers of DNN; $g_l(t) = ReLU(W_l^T t)$ and $ReLU(t) = max(0, t)$. It is important to note that $V_{ik} = V(z_i, \tilde x_i)$ implies that the utility of an alternative $k$ is the function of the attributes of \textit{all} the alternatives $\tilde x_i$ and the decision maker's socio-economic variables $z_i$. Equation \ref{eq:util_specification} illustrates that $V_{ik}$ becomes alternative-specific only in the final layer prior to the Softmax function when $w_k$ is applied to $\Phi(z_i, \tilde x_i)$. 

\subsection{Architecture of ASU-DNN}
\begin{figure}[ht]
\centering
\includegraphics[width=0.4\linewidth]{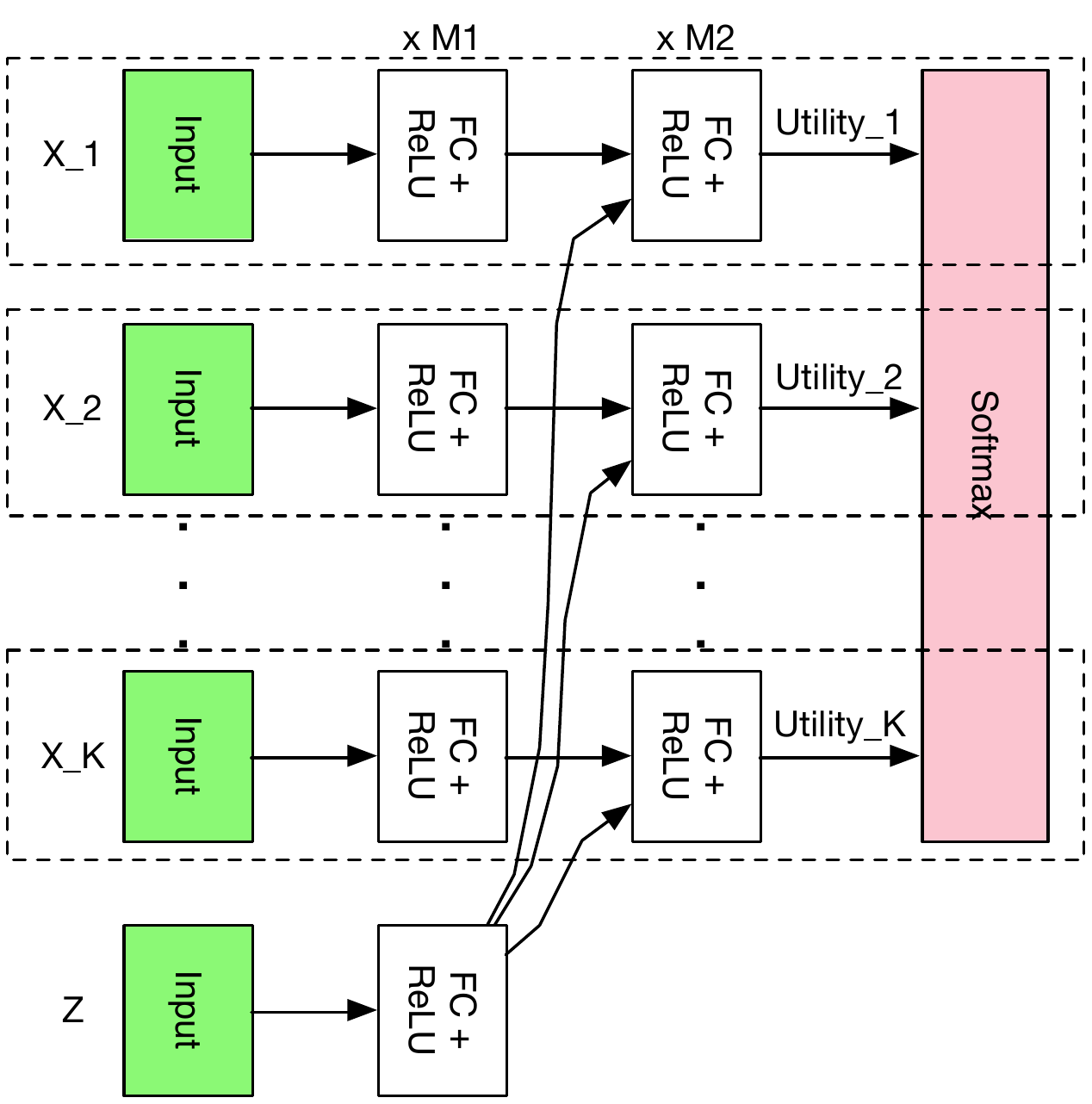}
\caption{ASU-DNN; Deep neural network architecture based on utility theory. It could be considered as a stack of fully connected subnetworks, with each computing a utility score for each alternative. Individual-specific variables interact with alternative-specific variables after $M_1$ layers.}
\label{fig:arch2}
\end{figure}

\noindent 
This utility insight enables us to design a DNN architecture with alternative-specific utility function, which is commonly assumed in choice models. Figure \ref{fig:arch2} shows the architecture of ASU-DNN. Herein, each alternative-specific $x_{ik}$ and individual-specific $z_i$ undergo transformation first, and $z_i$ enters the pathway of $x_{ik}$ after $M_1$ layers. As a result, the utility of each alternative becomes only a function of its own attributes $x_{ik}$ and of the decision maker's socio-demographic information $z_i$. This ASU-DNN dramatically reduces the complexity of F-DNN, while still capturing the heterogeneity of the utility function, which varies with the decision makers' socio-demographics. ASU-DNN could be considered as a stack of $K$ subnetworks, interacting with socio-demographics $z_i$. In addition, this alternative-specific utility is equivalent to the constraint of independence of irrelevant alternative (IIA) in this DNN setting. This is because the ratio of the choice probabilities of two alternatives no longer depends on other irrelevant alternatives. Formally, the utility function in ASU-DNN becomes
\begin{equation}
\setlength{\jot}{2pt} \label{eq:util_specification_2}
  \begin{aligned}
  V_{ik} &= V(z_i, x_{ik}) = w_k^T \Phi(z_i, x_{ik}) = w_k^T (g_{M_2} ... \circ g_2 \circ g_1)((g^{x_k}_{M_1} ... \circ g^{x_k}_1)(x_{ik}), (g^{z}_{M_1} ... \circ g^{z}_1)(z_i)) \\
  \end{aligned}
\end{equation}

This ASU-DNN architecture can potentially address the three challenges mentioned at the beginning of this work. First, this architecture is a compromise between domain-specific knowledge and a generic-purpose DNN model. On the one hand, the design permits only alternative-specific connectivity based on the utility theory, whereby the meta-architecture is handcrafted. On the other hand, the fully connected layers in ASU-DNN exploit the automated feature learning capacity of DNN. Therefore, the sub-network in ASU-DNN still uses the power of DNN as a universal approximator \cite{Cybenko1989,Hornik1989,Hornik1991}. Secondly, this alternative-specific connectivity design could provide more regular information than F-DNN owing to the underlying utility theory. The two architectures in Figures \ref{fig:arch1} and \ref{fig:arch2} are associated with different behavioral mechanisms. F-DNN implies that the utility of each alternative depends on the other alternatives. A good example is the reference-dependent utilities: when people use the market average price as a reference point, the utility of an alternative depends directly on other alternatives \cite{Weaver2009,Dhami2016}. Meanwhile, the baseline utility theory indicates that the utility of an alternative depends on only the attributes of that alternative. Hence the comparison between the two architectures could be considered as a test between two behavioral mechanisms. Thirdly, F-DNN has substantially more parameters than ASU-DNN does. When both the DNN architectures have $10$ layers and approximately $600$ neurons in each layer, F-DNN has approximate three million parameters, whereas ASU-DNN has $0.5$ million. Therefore, the alternative-specific connectivity design could be considered as a sparse architecture that regularizes DNN models. However, to formally evaluate the effectiveness of this regularization, the statistical learning theory is required to discuss the tradeoff between the approximation and estimation errors, as shown in the next section.

\subsection{Estimation and Approximation Error Tradeoff Between ASU-DNN and F-DNN}
\noindent
It is not true that ASU-DNN can always outperform F-DNN. This is because any constraint applied to DNN could potentially cause misspecification errors. Let ${\F}_1$ and ${\F}_2$ denote the model family of ASU-DNN and F-DNN; use $\hat{f}_1$ and $\hat{f}_2$ to denote the estimated decision rules from ASU-DNN and F-DNN, and $f^*$ to denote the true data generating process (DGP). The \textit{Excess error} is:
\begin{flalign}
{\E}_S [L(\hat f) - L(f^*)] &= {\E}_S[L(\hat f) - L(f^*_F)] + {\E}_S[L(f^*_F) - L(f^*)], \ \ \F \in \{ {\F}_1, {\F}_2 \}; \hat{f} \in \{ \hat{f}_1, \hat{f}_2 \}
\label{eq:excess_error_decomposition}
\end{flalign}

\noindent
where $L = {\E}_{x,y} [l(y, f(x)]$ is the expected loss function and $S$ represents the sample $\{x_i, y_i \}_1^N$. $f^*_{F} = \underset{f \in \F}{\argmin} \ L(f)$, the best function in function class $\F$ to approximate $f^*$. The excess error measures the average out-of-sample performance difference between the estimated function $\hat{f}$ and the true model $f^*$. The excess error can be decomposed as an \textit{estimation error}
\begin{flalign}
{\E}_S[L(\hat f) - L(f^*_F)]
\end{flalign}

\noindent
And an \textit{approximation error}
\begin{flalign}
{\E}_S[L(f^*_F) - L(f^*)]
\end{flalign}

Formally, the statistical learning theory could demonstrate that ASU-DNN outperforms F-DNN owing to the smaller estimation error of ASU-DNN. However, F-DNN could possibly outperform ASU-DNN owing to the smaller approximation error of F-DNN. When ASU-DNN and F-DNN have equal width and depth, the approximation error of ASU-DNN (${{\F}_1}$) is larger: 
\begin{flalign}
{\E}_S[L(f^*_{{\F}_1}) - L(f^*)] \geq {\E}_S[L(f^*_{{\F}_2}) - L(f^*)], \ \ {\F}_1 \subset {\F}_2
\end{flalign}

\noindent
This is intuitive because $f^*_{{\F}_1}$ also belongs to model family ${\F}_2$ and thus $f^*_{{\F}_2}$ could outperform $f^*_{{\F}_1}$ in terms of approximating the true model $f^*$. A more challenging question is regarding the estimation errors, the proof of which relies on the empirical process theory that uses Rademacher complexity as an upper bound. 

\begin{definition} \label{def:1}
Empirical Rademacher complexity of function class $\F$ is defined as:
\begin{flalign}
\hat{\R}_n({\F}|_S) = {\E}_{\epsilon} \ \underset{f \in {\F}}{\sup} \ \frac{1}{N} \sum_{i=1}^N \epsilon_i f(x_i) 
\end{flalign}
$\epsilon_i$ is the Rademacher random variable, taking values $\{-1, +1\}$ with equal probabilities.
\end{definition}

\begin{proposition} \label{prop:3}
\textit{The estimation error of an estimator $\hat{f}$ can be bounded by the Rademacher complexity of $\F$.}
\begin{flalign}
{\E}_S[L(\hat f) - L(f^*_F)] \leq 2{\E}_S \hat{\R}_n({\F}|_S) 
\end{flalign}
\end{proposition}

\noindent
Definition \ref{def:1} provides a measurement for the complexity of the function class $\F$. Proposition \ref{prop:3} implies that the estimation error is controlled by the complexity of $\F$. This is consistent with traditional wisdom that the estimation error increases when the number of parameters in a model is larger. Details of Definition \ref{def:1} and Proposition \ref{prop:3} are available in recent studies about the statistical learning theory \cite{Vershynin2018,Wainwright2019,Bartlett2002}.

\begin{proposition} \label{prop:4}
\textit{Let $H_d$ be the class of neural network with depth $D$ over the domain $\X$, where each parameter matrix $W_j$ has the Frobenius norm at most $M_F(j)$, and with ReLU activation functions. Then}
\begin{flalign}
\hat{\R}_n({\F}|_S) \leq \frac{(\sqrt{2 \log(D)} + 1) \sqrt{\frac{1}{N} \sum_{i=1}^N ||x_i||^2}}{\sqrt{N}} \times \prod_{j = 1}^D M_F(j) 
\end{flalign}
\end{proposition}

\noindent
Remarks on Proposition \ref{prop:4}:

\begin{enumerate}
\item As this result is from Golowich et al. (2017) \cite{Golowich2017}, so its proof is omitted in this study. Other relevant proofs are available in \cite{Bartlett2002,Neyshabur2015,Anthony2009}.
\item Proposition \ref{prop:4} indicates that the estimation error of DNN is a function of the depth $D$, Frobenius norm of each layer $M_F(j)$, diameter of $x$, and sample size $N$. 
\item Unlike traditional results based on VC-dimension \cite{Vapnik1999,Bartlett2017}, this upper bound relies on the norm of coefficients in each layer, which can be controlled by $l_1$ or $l_2$ regularizations, rather than the number of parameters. 
\item Suppose the width of DNN is $T$ and each entry in $W_j$ is at most $c$. The upper bound of F-DNN (${{\F}_2}$) in Proposition \ref{prop:4} can be re-expressed as:
\begin{flalign} \label{eq:estimation_error_f_dnn}
\hat{\R}_n({\F}_2|_S) \leq \frac{(\sqrt{2 \log(D)} + 1) \sqrt{\frac{1}{N} \sum_{i=1}^N ||x_i||^2}}{\sqrt{N}} \times c^D T^D
\end{flalign}
\end{enumerate}

\begin{proposition} \label{prop:5}
\textit{Suppose ASU-DNN has a total depth $D$ over the domain $\X$, wherein each entry in the matrix $W_j$ is at most $c$ and the width $T = K T_{x}$. $K$ is the number of alternatives in each choice scenario and $T_x$ is the width of each sub-network \footnote{This assumption simplies the ASU-DNN by omitting the socioeconomic inputs, because adding socioeconomic inputs into this proposition does not change our main conclusion.}. With ReLU activation functions}
\begin{flalign} \label{eq:estimation_error_asu_dnn_1}
\hat{\R}_n({\F}_1|_S) \leq \frac{(\sqrt{2 \log(D)} + 1) \sqrt{\frac{1}{N} \sum_{i=1}^N ||x_i||^2}}{\sqrt{N}} \times \frac{c^D T^D}{K^{D/2}}
\end{flalign}
\end{proposition}

\noindent
Remarks on Proposition \ref{prop:5}:

\begin{enumerate}
\item Proposition \ref{prop:5} can be derived from Proposition \ref{prop:4} by plugging in the coefficient matrix of each layer in ASU-DNN. 
\item The estimation error of ASU-DNN (${\F}_1$) shrinks by a factor of $O(K^{D/2})$ compared to F-DNN (${\F}_2$), implying that ASU-DNN performs better than F-DNN as $K$ or $D$ increases. 
\end{enumerate}

Equations \ref{eq:excess_error_decomposition}-\ref{eq:estimation_error_asu_dnn_1} constitute the formal method for illustrating the tradeoff between ASU-DNN and F-DNN. Owing to its sparse connectivity, ASU-DNN has smaller estimation error as its main advantage, particularly when K is large, as shown in Equation \ref{eq:estimation_error_asu_dnn_1}. Meanwhile, the larger approximation error could be the main disadvantage of ASU-DNN. When the alternative-specific utility constraint is not true in reality, this constraint could be excessively restrictive, resulting in a low model performance. This problem is also commonly acknowledged in the field of choice modeling, although framed in a different way. Because the alternative-specific utility function in this DNN setting indicates the IIA constraint, the large approximation error of ASU-DNN could be equivalently framed as a problem of IIA being too restrictive. This drawback appears unavoidable in the approach wherein DNN’s interpretability is improved ex-ante. This is because any prior knowledge may be too restrictive in reality. However, compared to classical choice modeling methods that rely exclusively on handcrafted feature learning, misspecification in ASU-DNN is less problematic because it is robust to utility specification \textit{conditioning on} the alternative-specific utility constraint. In addition, Equations \ref{eq:estimation_error_f_dnn} and \ref{eq:estimation_error_asu_dnn_1} indicate that the estimation error gap between ASU-DNN and F-DNN could reduce as the sample size increases. Overall, the trade-off between ASU-DNN and F-DNN involves complex dynamics between true models, sample size, number of alternatives, and regularization strength. To compare their performance, we need to apply them to real choice datasets.

\section{Setup of Experiments}
\subsection{Datasets}
\noindent
Our experiments are based on two datasets, an online survey data collected in Singapore with the aid of a professional survey company and a public dataset in R mlogit package. They are referred to as SGP and TRAIN, respectively, in this study. The SGP survey consisted of a section of choice preference and a section for eliciting socioeconomic variables. At the beginning, all the respondents reported their home and working locations and present travel mode. After obtaining the geographical information, our algorithm computed the walking time, waiting time, in-vehicle travel time, and travel cost of each travel mode based on the origin and destination provided by the participants and the price information collected from official data sources in Singapore. The SGP and TRAIN datasets include $8,418$ and $2,929$ observations. In the SGP dataset, the output $y_i$ represents the travel mode choice among walking, public transit, driving, ride sharing, and autonomous vehicles (AV); alternative-specific inputs $x_{ik}$ are the attributes of each travel mode, such as price and time cost; and individual-specific inputs $z_i$ are the attributes of decision-makers, such as their income and education backgrounds. In the TRAIN dataset, $y_i$ represents the binary travel mode choice between two different types of trains; the alternative-specific input $x_{ik}$ represents the price, time cost, and level of comfort; and no $z_i$ exists for the TRAIN dataset. Both of the datasets are divided into training, validation, and testing sets in the ratio $4:1:1$. Five-fold cross-validation is used for the model selection, and the model evaluation is based on both the validation and testing sets. Detailed summary statistics of TRAIN and SGP are attached in Appendix II.

\subsection{Hyperparameter Space}
\noindent
A challenge in the comparison between the two DNN architectures is the large number of hyperparameters, on which the performance of DNN largely depends. Table \ref{table:arch_hpo_1} summarizes a list of hyperparameters and the range of their values. The hyperparameters consist of invariant ones, varying ones specific to F-DNN or ASU-DNN, and varying ones shared by F-DNN and ASU-DNN. The difference between F-DNN and ASU-DNN is referred to as alternative-specific connectivity hyperparameter, which plays a similar role as the other hyperparameters do because it changes the architecture of DNN, controls the number of parameters, and performs regularization.

\begin{table}[ht]
\caption{Hyperparameter space of F-DNN and ASU-DNN; \textit{Panel 1}. Hyperparameters that don't change in the hyperparameter searching; \textit{Panel 2}. Hyperparameters that change in only F-DNN; \textit{Panel 3}. Hyperparameters that change in only ASU-DNN; $M_1$ and $n_1$ are the depth and width before the interaction between $x_{ik}$ and $z_i$. \textit{Panel 4}. Hyperparameters that change in both F-DNN and ASU-DNN.}
\centering
\resizebox{0.9\linewidth}{!}{%
\begin{tabular}{P{0.3\linewidth} | P{0.6\linewidth}}
\hline
\textbf{Hyperparameters} & \textbf{Values} \\
\hline
\hline
\multicolumn{2}{l}{\textit{Panel 1. Invariant Hyperparameters}} \\
\hline
Activation functions & ReLU and Softmax \\
\hline
Loss & Cross-entropy \\
\hline
Initialization & He initialization \\
\hline
\hline
\multicolumn{2}{l}{\textit{Panel 2. Varying Hyperparameters of F-DNN}} \\
\hline
M & $[1,2,3,4,5,6,7,8,9,10,11,12]$ \\
\hline
Width $n$ & $[60, 120, 240, 360, 480, 600]$ \\
\hline
\hline
\multicolumn{2}{l}{\textit{Panel 3. Varying Hyperparameters of ASU-DNN}} \\
\hline
$M_1$ & $[0,1,2,3,4,5,6]$ \\
\hline
$M_2$ & $[0,1,2,3,4,5,6]$ \\
\hline
Width $n_1$ & $[10, 20, 40, 60, 80]$ \\
\hline
Width $n_2$ & $[10, 20, 40, 60, 80, 100]$ \\
\hline
\hline
\multicolumn{2}{l}{\textit{Panel 4. Varying Hyperparameters of F-DNN and ASU-DNN}} \\
\hline
$\gamma_{1}$ ($l_1$ penalty) & $[1.0, 0.5, 0.1, 0.01, 10^{-3}, 10^{-5}, 10^{-10}, 10^{-20}]$ \\
\hline
$\gamma_{2}$ ($l_2$ penalty) & $[1.0, 0.5, 0.1, 0.01, 10^{-3}, 10^{-5}, 10^{-10}, 10^{-20}]$ \\
\hline
Dropout rate & $[0.5, 0.1, 0.01, 10^{-3}, 10^{-5}]$ \\
\hline
Batch normalization & $[True, False]$ \\
\hline
Learning rate & $[0.5, 0.1, 0.01, 10^{-3}, 10^{-5}]$ \\
\hline
Num of iteration & $[500, 1000, 5000, 10000, 20000]$ \\
\hline
Mini-batch size & $[50, 100, 200, 500, 1000]$ \\
\hline
\end{tabular}
} 
\label{table:arch_hpo_1}
\end{table}

A brief introduction for some hyperparameters is as following. \textbf{Activation Functions.} Rectified linear unit (ReLU) is used in the middle layers and Softmax is used in the last layer. Other activation functions are also possible, although recent studies have shown that non-saturated activation functions (e.g. ReLU) perform better than the saturated activation functions (e.g. Tanh) \cite{Krizhevsky2012}. \textbf{Initialization.} It refers to the process of initializing the parameters in DNN. DNN initialization does not have formal theory yet, although Glorot and He initializations are commonly used in practice \cite{Glorot2010,Glorot2011,HeKaiming2015_2}. \textbf{Depth and Width.} They refer to the number of layers and the number of neurons in each layer of DNN. Depth and width control the model complexity: DNN models have smaller approximation errors and larger estimation errors, when they become wider and deeper. \textbf{Penalties.} Both $l_1$ and $l_2$ penalties are explicit regularization added to the standard cross-entropy loss function. The $l_1$ penalty encourages model sparsity; the $l_2$ penalty shrinks the magnitude of coefficients. \textbf{Dropout.} It refers to the process of randomly dropping certain proportion of the neurons in training \cite{Hinton2012}, and since this procedure leads to sparser architecture, it can also be treated as a regularization method. \textbf{Batch Normalization.} It is the normalization of each batch in the stochastic gradient descent (SGD). \textbf{Number of Iterations.} It refers to the number of iterations in the training. Too few training iterations could lead to an underfitted model and too many iterations could lead to an overfitted model. As a result, a relatively small number of iterations (e.g. early stopping) can be considered as a regularization method.

\subsection{Hyperparameter Searching}
\noindent
It is a benchmark method to randomly search in the hyperparameter space to identify the DNN configuration with a high prediction accuracy \cite{Bergstra2012}. In our study, $100$ DNN models were trained, $50$ each for the two DNN architectures. Formally, the empirical risk minimization (ERM) is

\begin{equation}
\setlength{\jot}{2pt} \label{eq:erm}
  \begin{aligned}
  \underset{w}{\min} \ E(w, w_h) = \underset{w}{\min} \frac{1}{N} \sum_i^N l(y_i, P_{ik}; w, w_h) + \gamma ||w||_p
  \end{aligned}
\end{equation}

\noindent
in which $w$ represents parameters; $w_h$ represents hyperparameters; $l()$ is the cross-entropy loss function, and $\gamma ||w||_p$ represents $l_p$ penalty. Suppose $w^*$ minimizes $E(w, w_h)$ conditioning on one specific $w_h$. By randomly sampling $w_h^{(s)}$, we could identify the best hyperparameter $w_h^*$

\begin{equation}
\setlength{\jot}{2pt} \label{eq:hyper_searching_2}
  \begin{aligned}
  w_h^* = \underset{w_h \in \{w_h^{(1)},w_h^{(2)}, ..., w_h^{(S)} \} }{\argmin} E(w^*, w_h)
  \end{aligned}
\end{equation}

\section{Experiment Results}
\label{s:6}
\noindent
The result section consists of three parts. The first part compares the prediction accuracy of ASU-DNN, F-DNN, MNL, NL, and other nine ML classifiers. The second part evaluates how effective the alternative-specific connectivity is as a regularization method, as opposed to other generic-purpose regularization methods. The final part compares ASU-DNN, F-DNN, MNL, and NL in terms of their interpretability by visualizing their choice probability functions and computing their elasticity coefficients. The first part uses both SGP and TRAIN datasets, and the second and third parts focus on only the SGP dataset for simplicity.

\subsection{Prediction Accuracy}
\begin{figure}[t!]
\centering
\subfloat[SGP Validation]{\includegraphics[width=0.3\linewidth]{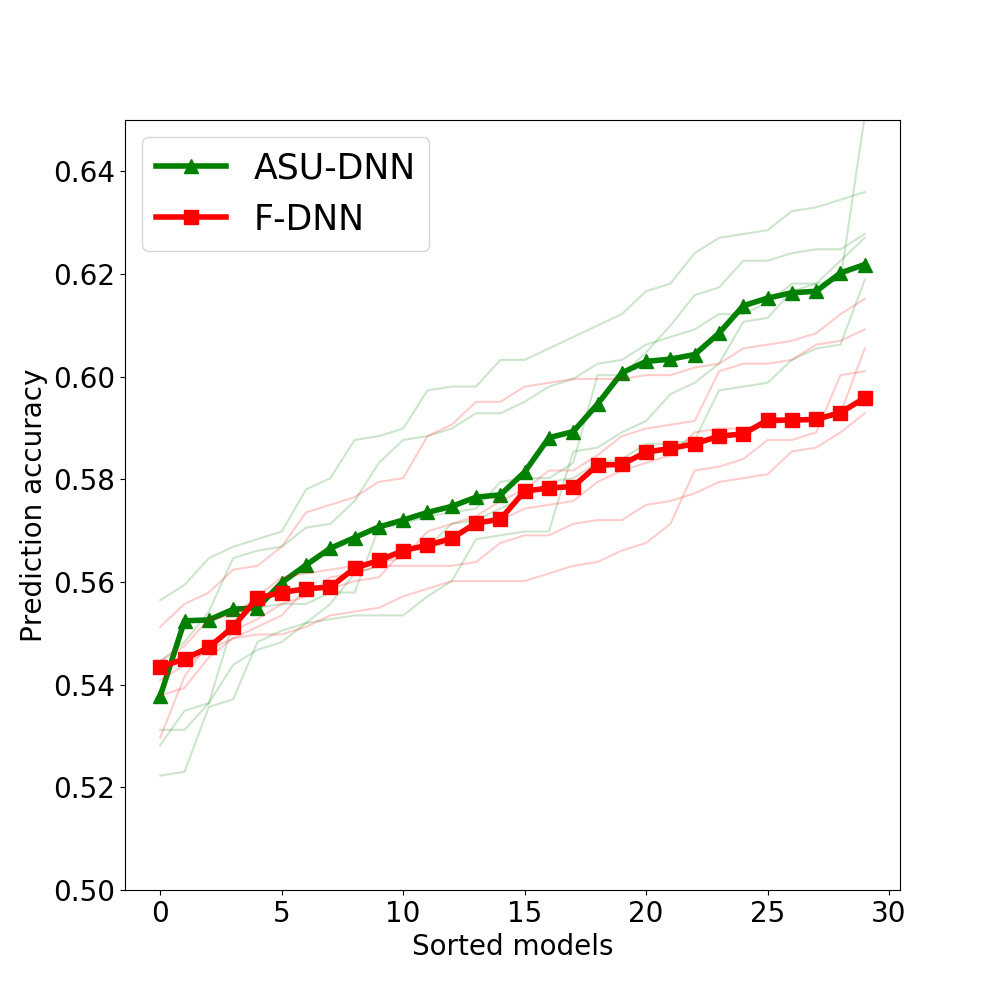}\label{sfig:hpo_val_sgp}} 
\subfloat[SGP Testing]{\includegraphics[width=0.3\linewidth]{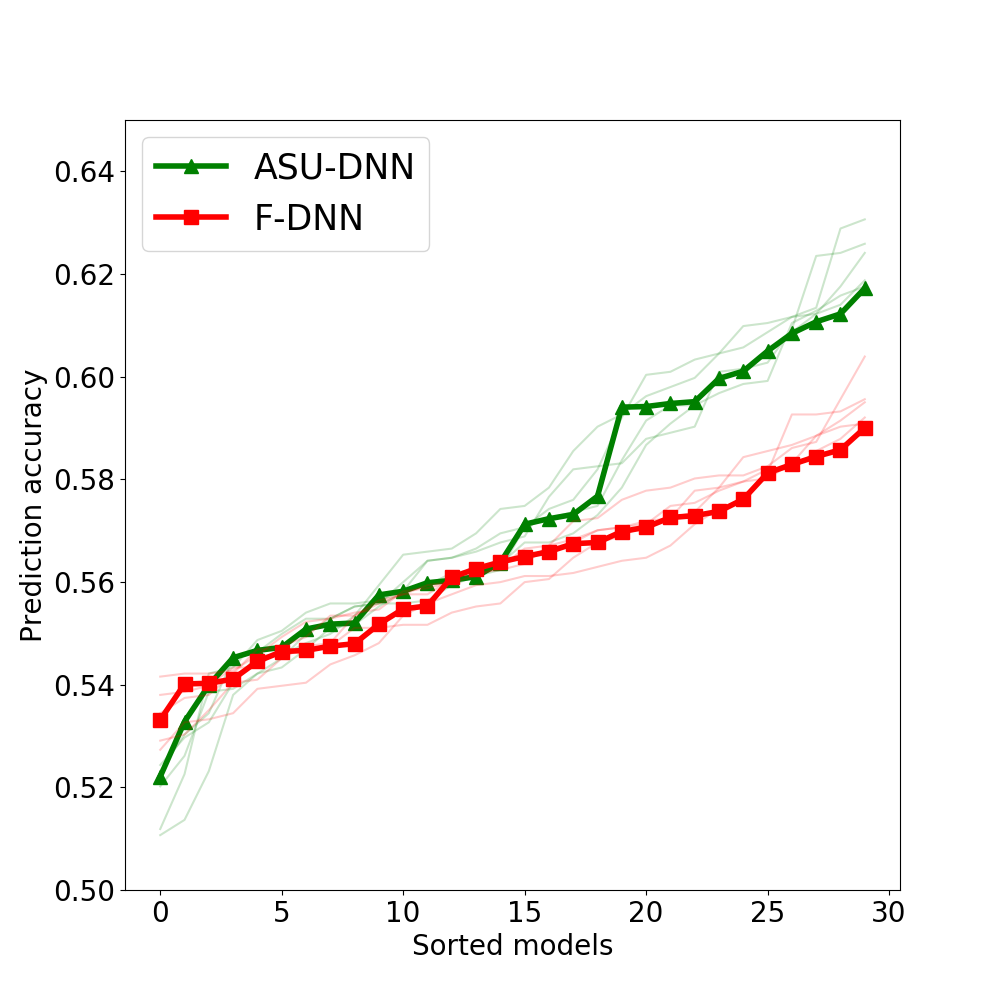}\label{sfig:hpo_test_sgp}}\\
\subfloat[TRAIN Validation]{\includegraphics[width=0.3\linewidth]{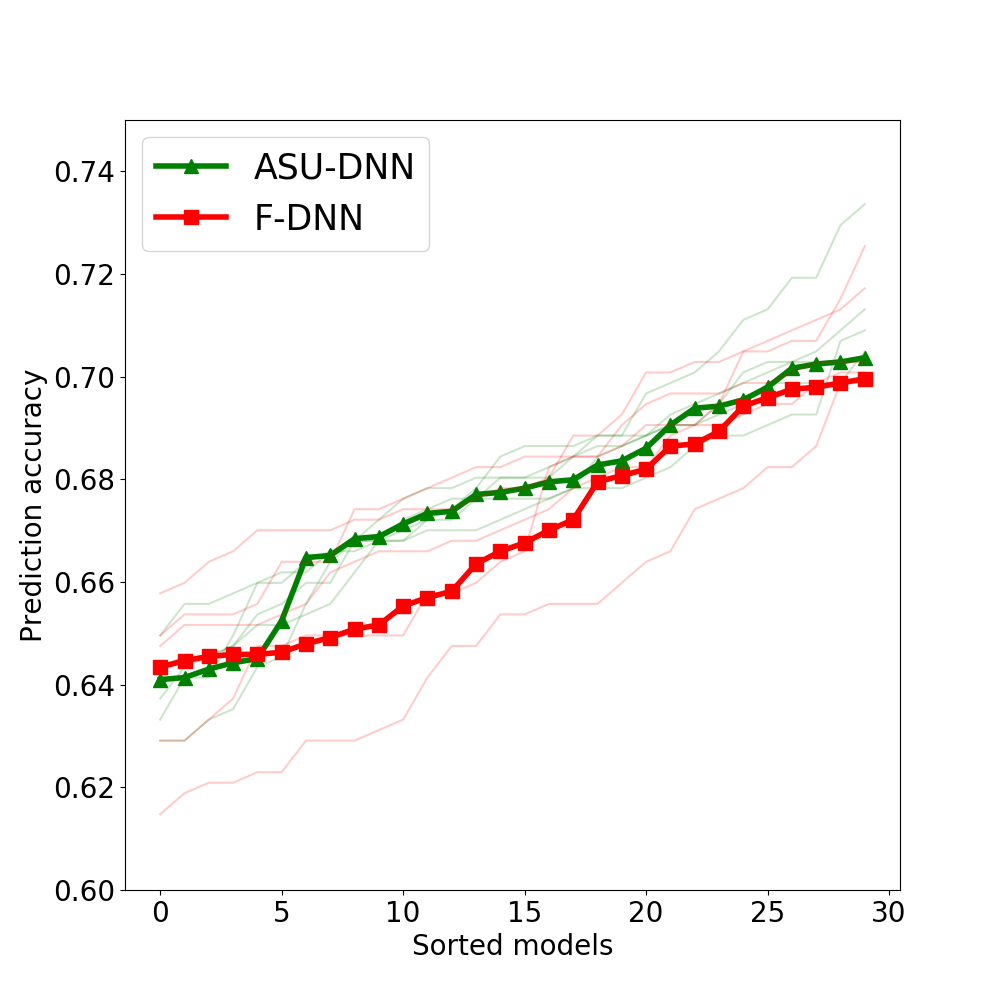}\label{sfig:hpo_val_train}} 
\subfloat[TRAIN Testing]{\includegraphics[width=0.3\linewidth]{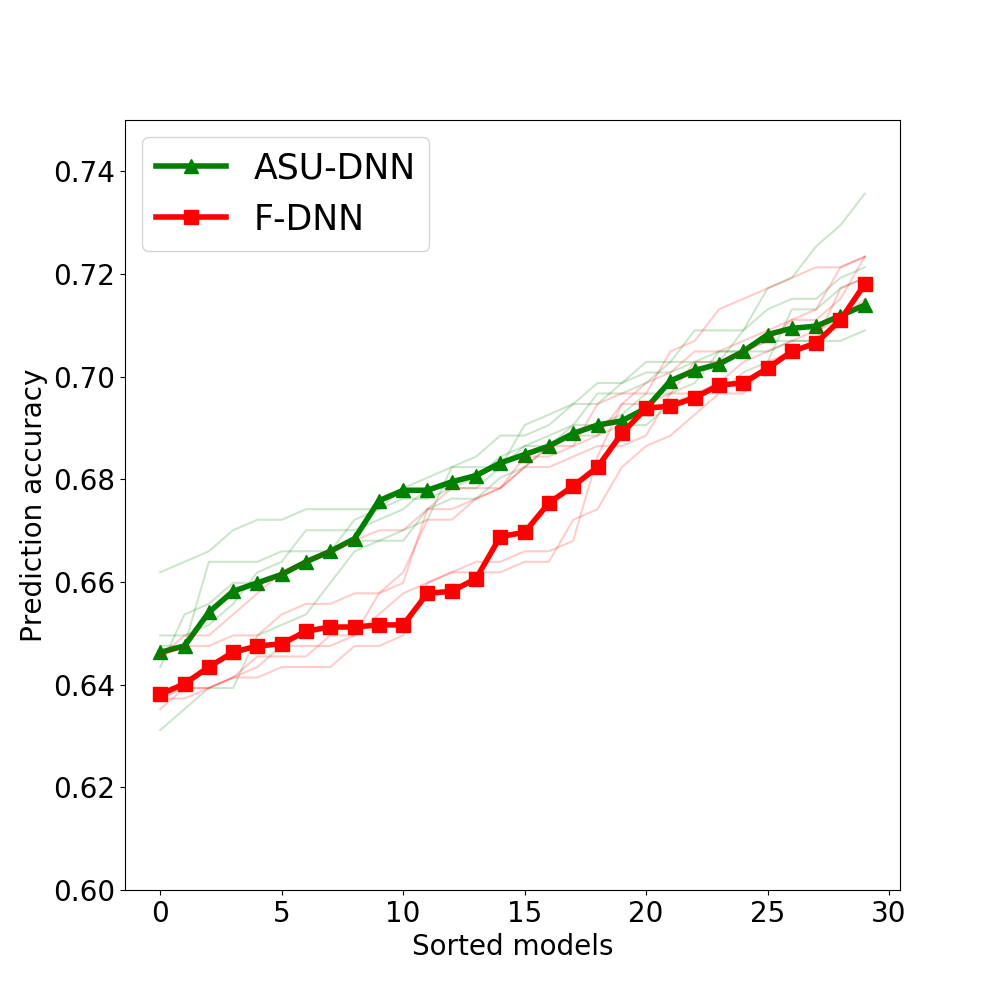}\label{sfig:hpo_test_train}}\\
\caption{Hyperparameter Searching Results; in all four subfigures, models are sorted according to prediction accuracy. Green curves represent ASU-DNN performance, and red ones represent F-DNN. Dark curves are the average of five-fold cross-validation, and light ones are the individual trainings. Overall, ASU-DNN consistently outperforms F-DNN. The information of top DNN architectures is attached in Appendix III.}
\label{fig:hpo}
\end{figure}

\noindent
Figure \ref{fig:hpo} summarizes the prediction accuracy of the top $30$ models in the validation and the testing sets in the SGP and TRAIN datasets. All the four figures illustrate that ASU-DNN performs better than F-DNN does, although there are marginal differences between the SGP and TRAIN datasets \footnote{Here we focus on only the top models since researchers only choose the top ML models for analysis. For example, researchers compare the top 1 model or the top 5 models in two different model families, so we don't discuss the mean or the variance of the models' performance.}. In the SGP dataset, the prediction accuracy of ASU-DNN in the first $15$ out of the visualized $30$ models is approximately $0.5\%$ higher than that of F-DNN. Moreover, the difference in prediction accuracy increases as the models' prediction accuracy increases. The top $10$ ASU-DNNs outperform the top $10$ F-DNNs by approximately $2-3 \%$ prediction accuracy in both validation and testing sets. The best ASU-DNN outperforms the best F-DNN by approximately $3\%$. In the TRAIN dataset, whereas the ASU-DNN still consistently outperforms F-DNN, the gap is smaller in its top $10$ models. The first $15$ out of the visualized $30$ ASU-DNN models outperform the F-DNN models by $2-3 \%$ of prediction accuracy, whereas the top $10$ ASU-DNNs outperform F-DNN by only $0.5\%$. An outlier case is the top $1$ model in the testing set of TRAIN; herein, the prediction accuracy of F-DNN is marginally higher than that of ASU-DNN. Nonetheless, it is evident that in nearly all the cases, ASU-DNN consistently performs higher than F-DNN does in the whole hyperparameter space. 

Table \ref{table:classifiers} also illustrates that both F-DNN and ASU-DNN perform better than the other eleven classifiers, implying that DNN models fit choice analysis tasks very effectively. Specifically, F-DNN and ASU-DNN outperform the baseline MNL and NL by about 8\% prediction accuracy, implying that the compositional function structure of DNN is effective. Because the prediction accuracy gap between ASU-DNN and F-DNN is identified by using random sampling from the hyperparameter space, we could attribute this gain in prediction accuracy to only the alternative-specific connectivity design and not to any other regularization method. In addition, from the perspective of the behavioral test, the better performance of ASU-DNN than F-DNN indicates that the utility of an alternative was computed based on its own attributes rather than the attributes of all the alternatives.

\begin{table}[htb]
\caption{Prediction accuracy of all classifiers; MNL represents the multinomial logit model and NL represents the nested logit model. Nest 1: walking + bus (the corresponding scale parameter $\mu_1$ is fixed to 1); Nest 2: AV + ridesharing + driving. NL is not applicable (N.A.) to the TRAIN data set because it has only two alternatives. LR (l1\_reg/l2\_reg) represents a logistic regression model with mild l1 or l2 regularization; SVM (Linear/RBF) represents for support vector machine with linear or RBF kernels; KNN\_3 represents three-nearest neighbor classifier; decision tree is abbreviated as DT; quadratic discriminant analysis is as QDA. The DNN models outperform all the other classifiers.}
    \centering
    \resizebox{1.00\linewidth}{!}{%
    \begin{tabular}{p{0.1\linewidth}|P{0.07\linewidth}|P{0.07\linewidth}|P{0.07\linewidth}|P{0.07\linewidth}|P{0.07\linewidth}|P{0.07\linewidth}|P{0.07\linewidth}|P{0.07\linewidth}|P{0.07\linewidth}|P{0.07\linewidth}|P{0.07\linewidth}|P{0.07\linewidth}|P{0.07\linewidth}|P{0.07\linewidth}|P{0.07\linewidth}}
        \toprule
         & ASU-DNN (Top 1) & F-DNN (Top 1) & ASU-DNN (Top 10) & F-DNN (Top 10) & MNL & NL & LR (l1\_reg) & LR (l2\_reg) & SVM (Linear) & SVM (RBF) & Naive Bayesian & KNN\_3 & DT & AdaBoost & QDA \\
        \midrule
        Validation (SGP) & 62.3\% & 59.2\% & 61.3\% & 58.8\% & 53.0\% & 54.1\% & 54.5\% & 54.7\% & 54.3\% & 45.6\% & 44.7\% & 58.5\% & 51.9\% & 54.6\% & 47.2\% \\
        \hline
        Test (SGP) & 61.0\% & 58.7\% & 60.4\% & 57.6\%  & 51.2\%  & 52.1\% & 52.1\% & 52.1\% & 51.8\% & 44.3\% & 41.6\% & 57.9\% & 50.2\% & 52.1\% & 44.9\% \\
        \hline
        Validation (TRAIN) & 70.5\% & 70.1\% & 69.8\% & 69.4\% & 69.4\% & N.A. & 69.5\% & 69.5\% & 68.8\% & 60.9\% & 57.3\% & 60.0\% & 65.0\% & 67.5\% & 60.2\% \\
        \hline
        Test (TRAIN) & 71.4\% & 72.1\% & 71.2\% & 70.7\% & 67.9\% & N.A. & 67.8\% & 67.9\% & 68.3\% & 58.7\% & 56.4\% & 57.7\% & 65.0\% & 69.8\% & 60.5\% \\
        \bottomrule
    \end{tabular}
    } 
    \label{table:classifiers}
\end{table}

\subsection{Alternative-Specific Connectivity Design and Other Regularizations}
\begin{figure*}[t!]
\subfloat[$l_1$ Regularization]{\includegraphics[width=0.25\linewidth]{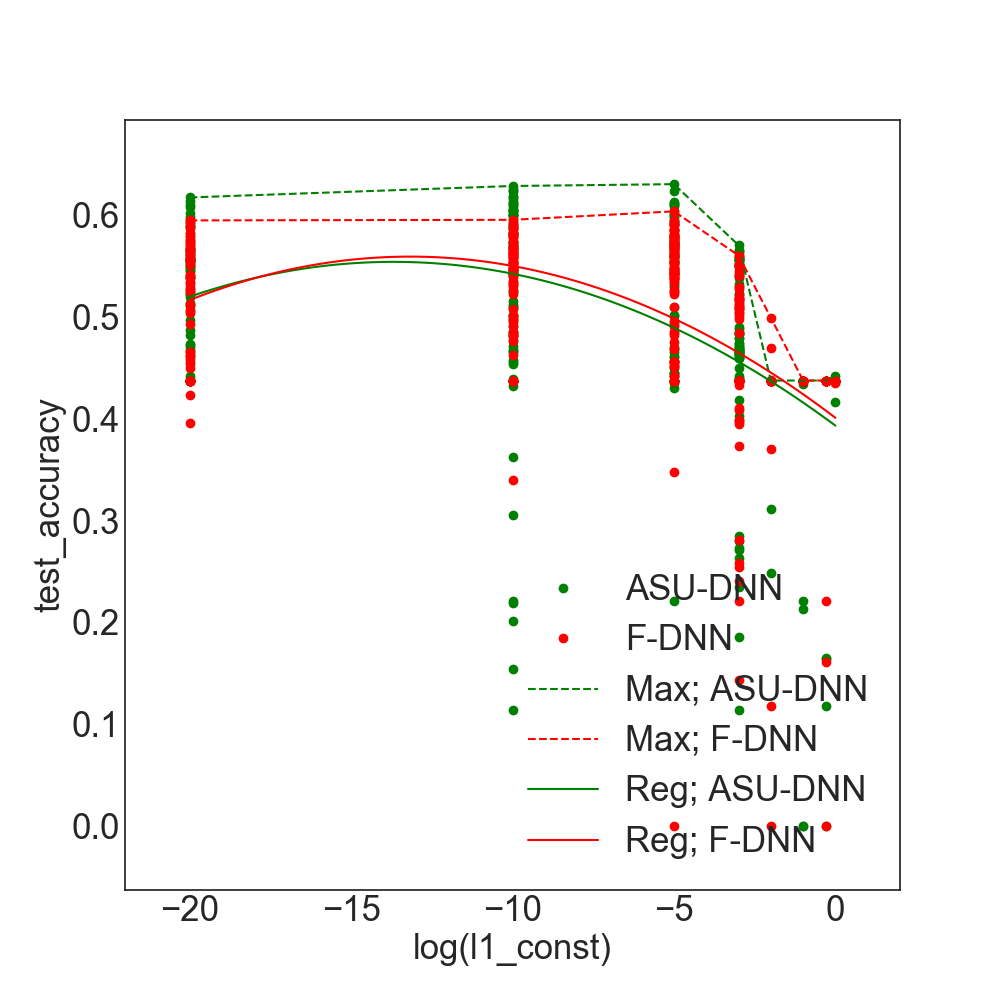}\label{sfig:l1_test}}
\subfloat[$l_2$ Regularization]{\includegraphics[width=0.25\linewidth]{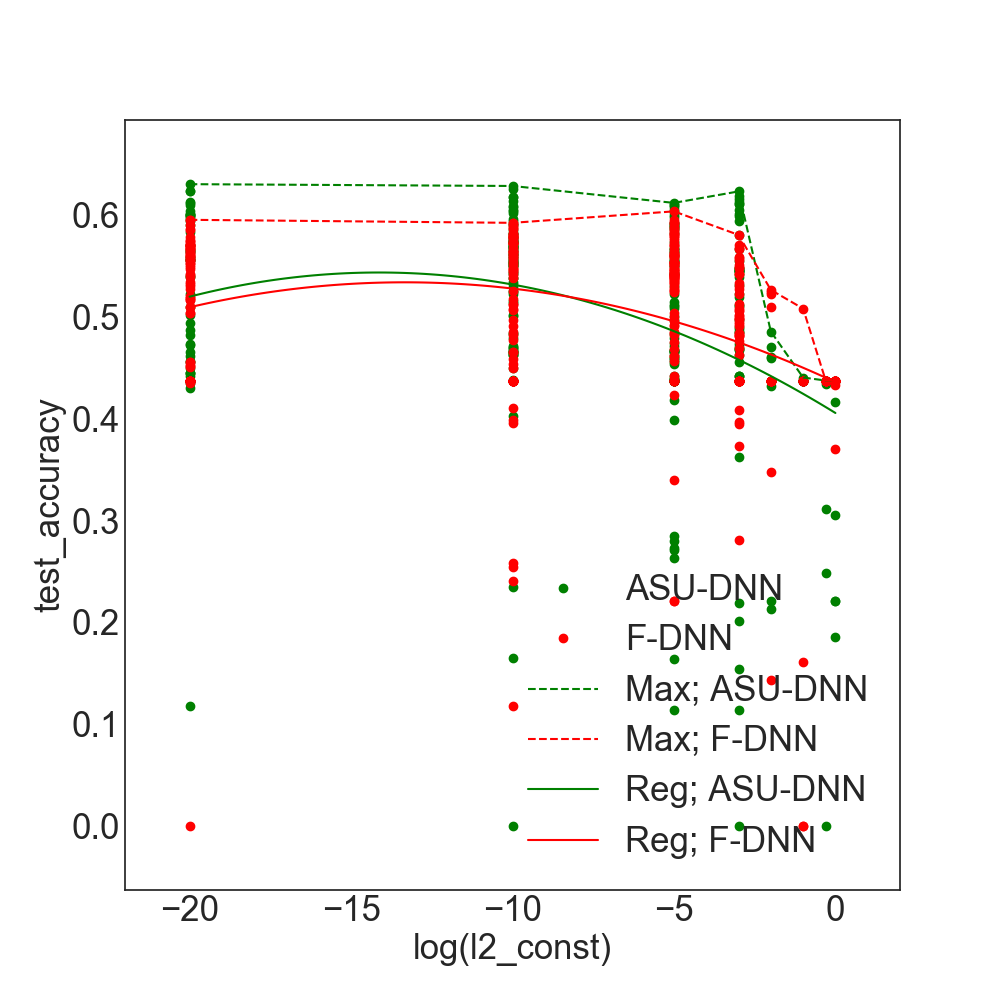}\label{sfig:l2_test}} \\
\subfloat[Learning Rates]{\includegraphics[width=0.25\linewidth]{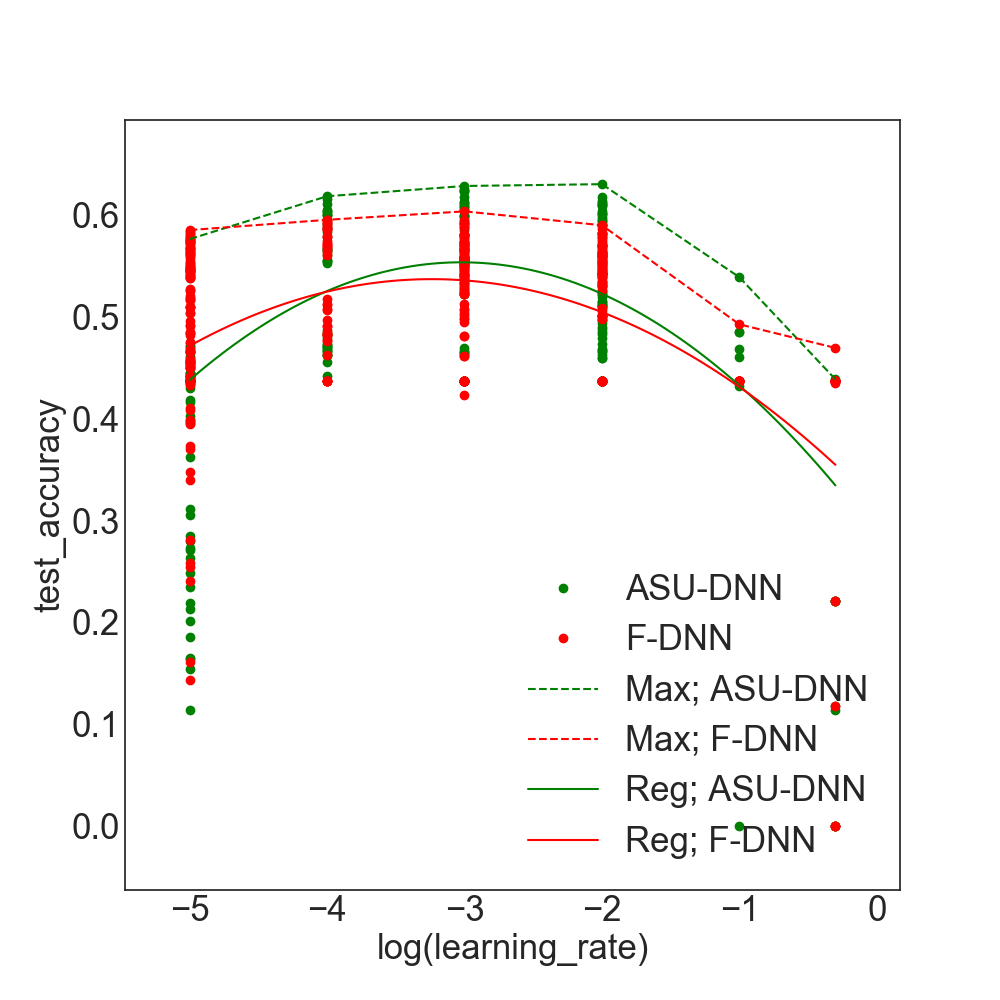}\label{sfig:lr_test}}
\subfloat[Number of Iteration]{\includegraphics[width=0.25\linewidth]{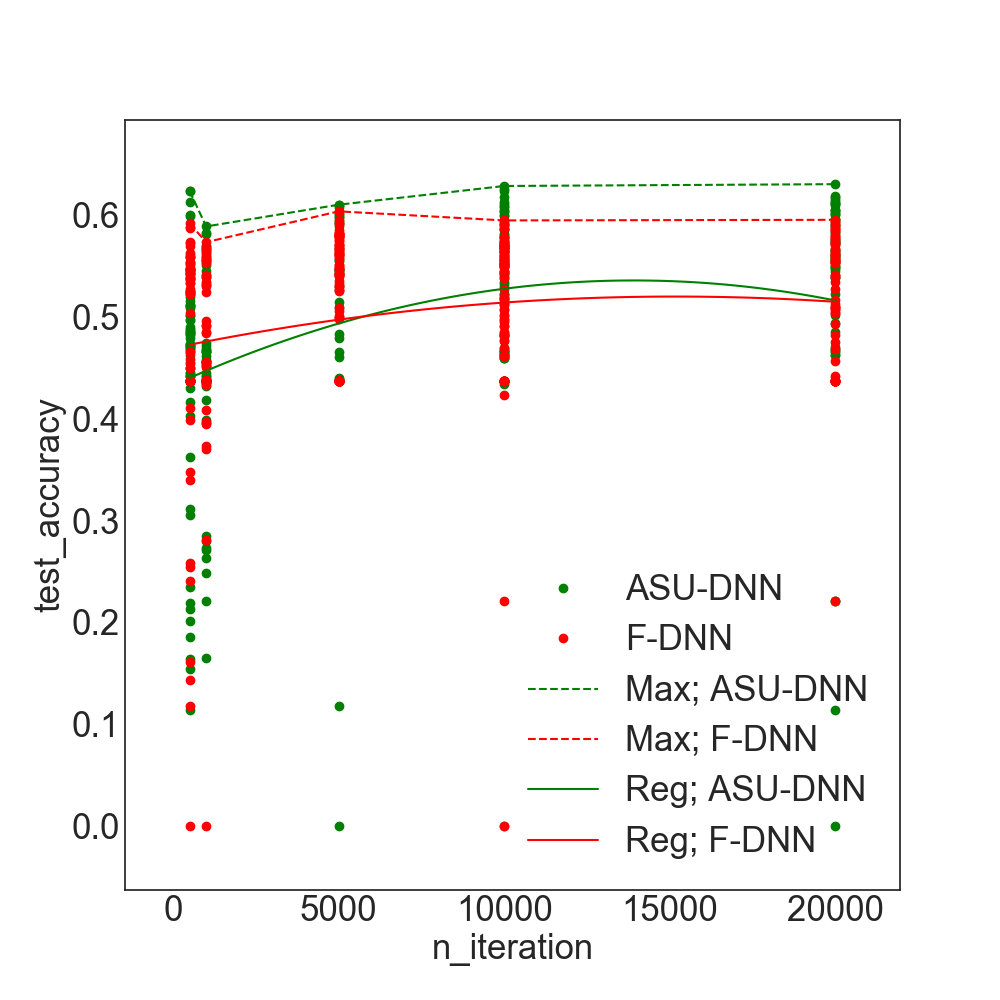}\label{sfig:ni_test}}
\subfloat[Size of Mini Batch]{\includegraphics[width=0.25\linewidth]{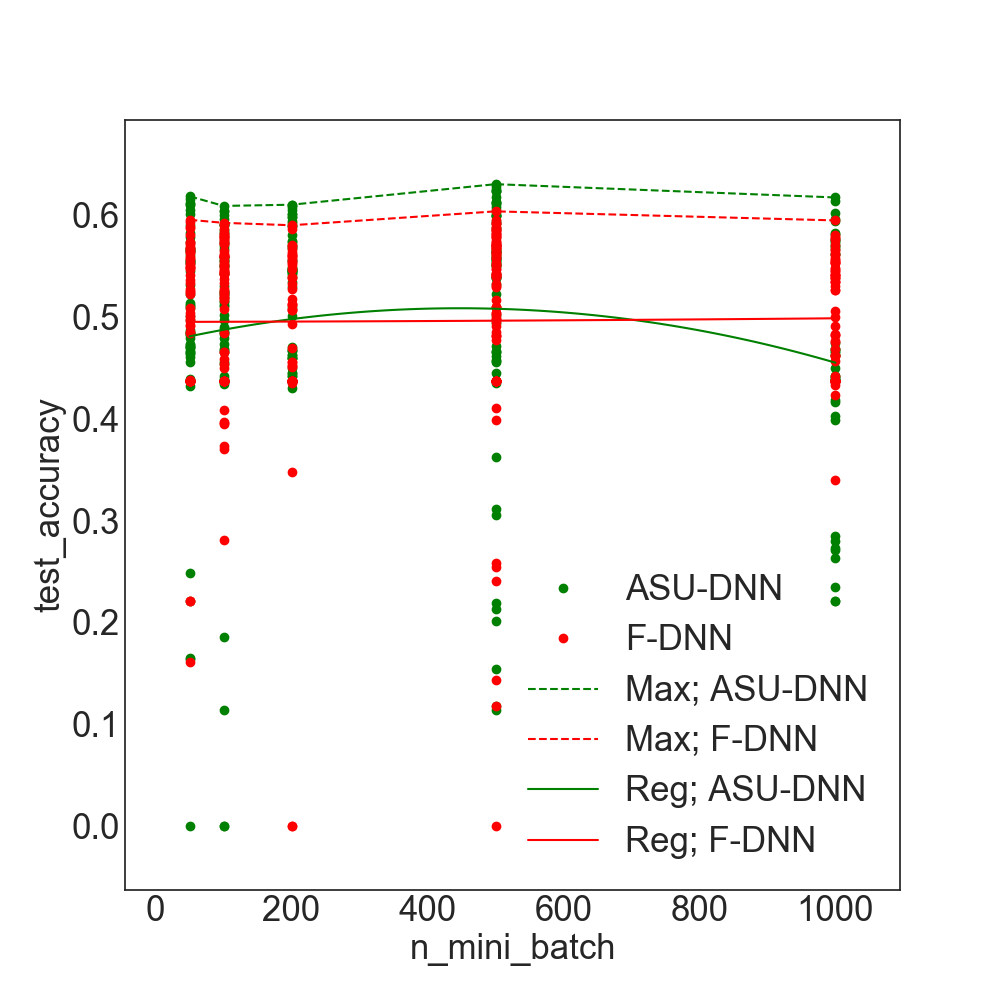}\label{sfig:n_mini_test}}
\subfloat[Batch Normalization]{\includegraphics[width=0.25\linewidth]{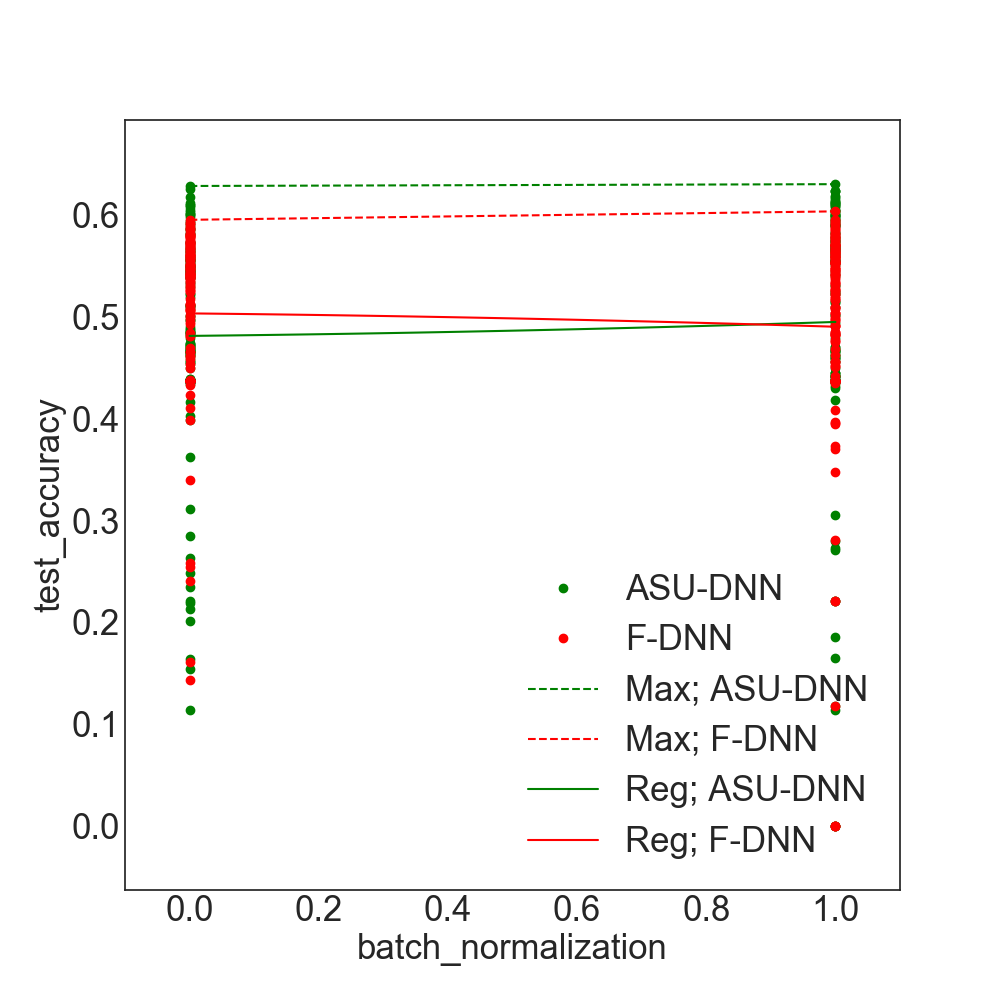}\label{sfig:bn_test}}\\
\subfloat[Depth of DNN]{\includegraphics[width=0.25\linewidth]{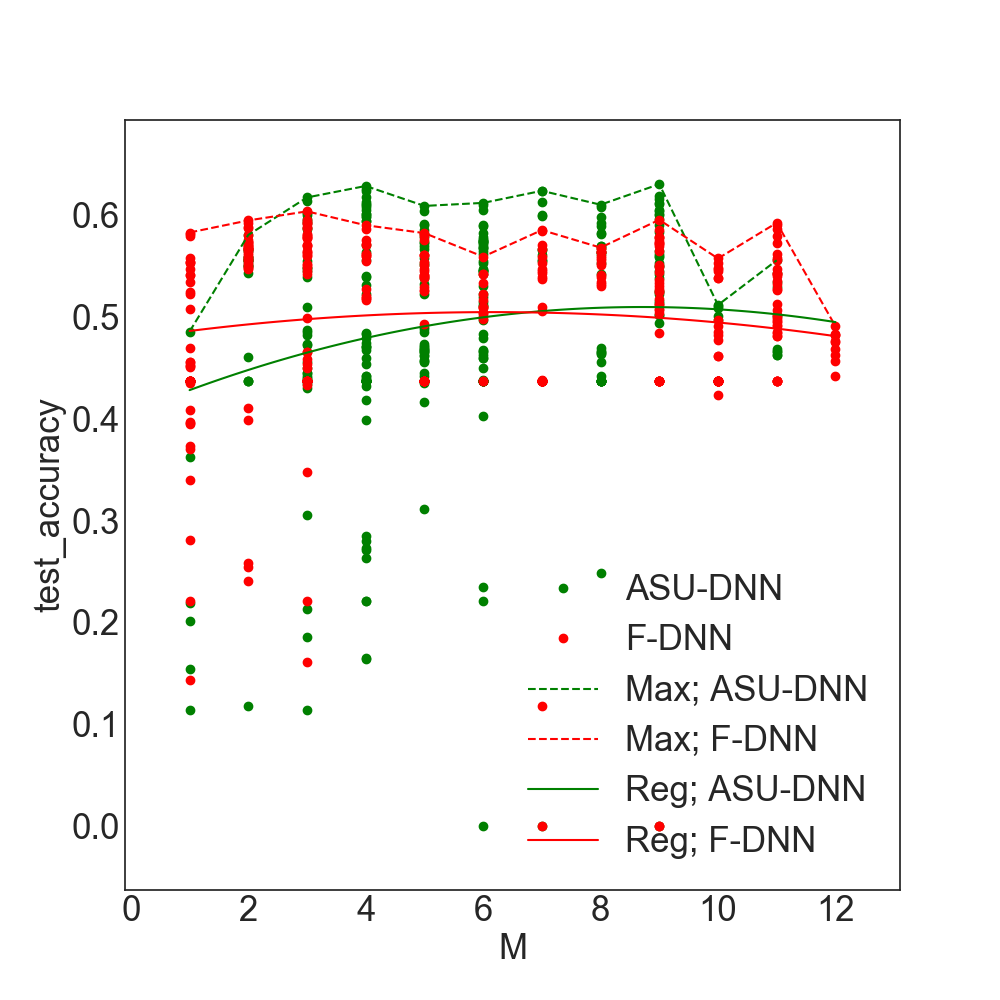}\label{sfig:M_test}}
\subfloat[Width of DNN]{\includegraphics[width=0.25\linewidth]{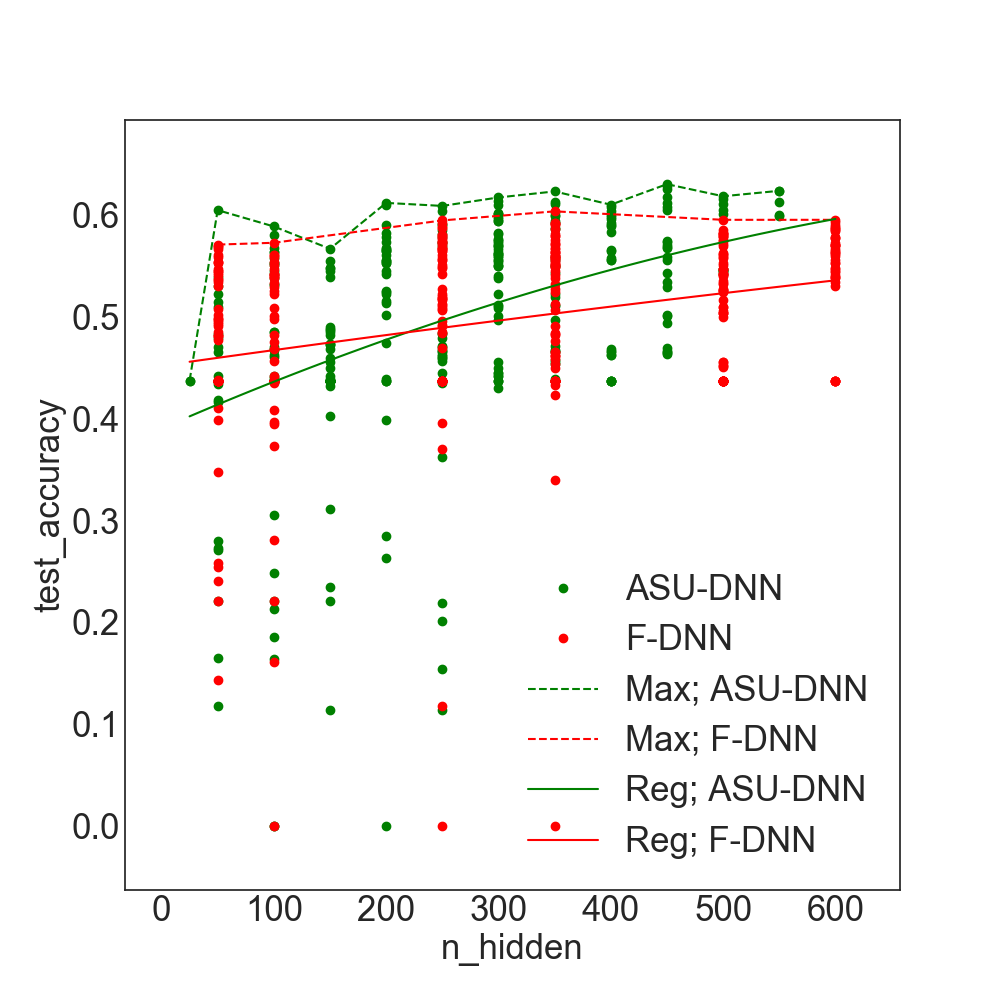}\label{sfig:n_hidden_test}}
\subfloat[Dropout Rates]{\includegraphics[width=0.25\linewidth]{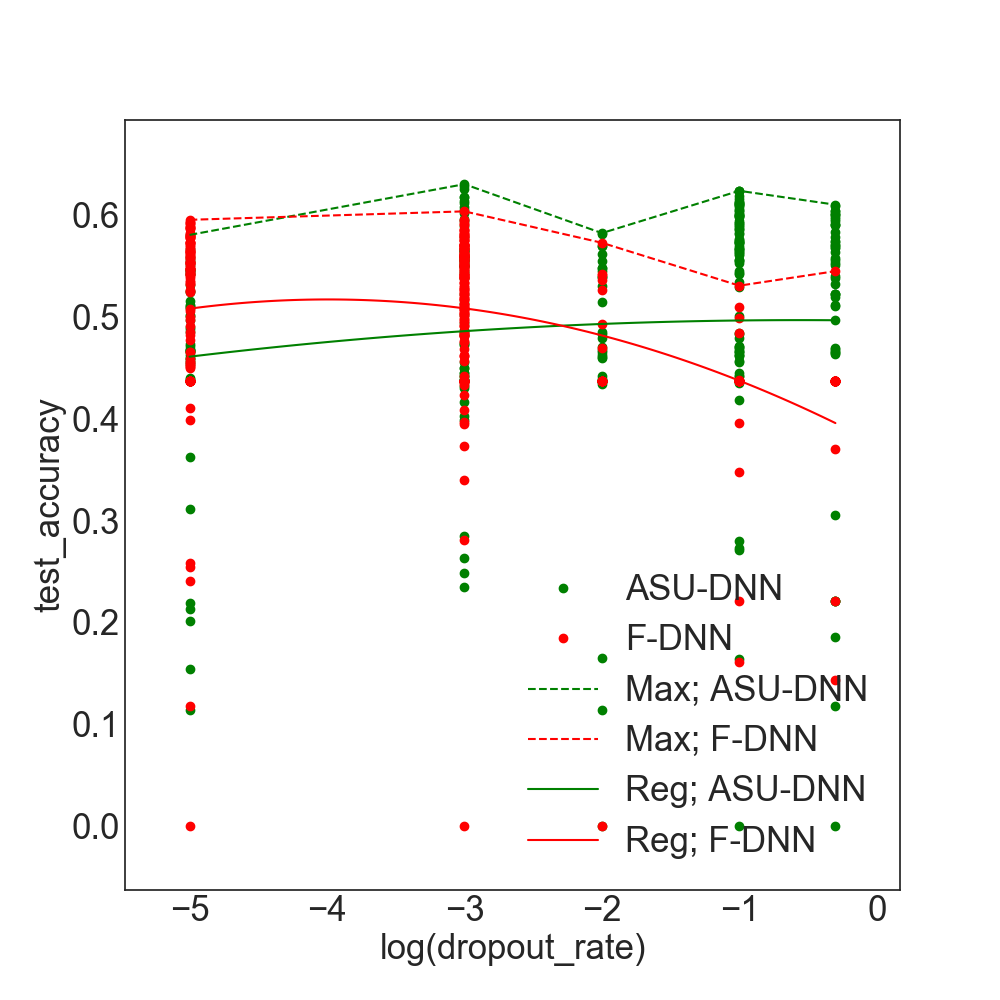}\label{sfig:dr_test}} \\

\caption{Comparing alternative-specific connectivity to explicit regularizations, implicit regularizations, and architectural hyperparameters in the SGP testing dataset; \textit{First row}: Explicit regularizations; \textit{Second row}: Implicit regularizations; \textit{Third row}: Architectural hyperparameters. In all the subfigures, the x-axis represents the hyperparameter and the y-axis represents the prediction accuracy. The dashed lines connect the models with the highest prediction accuracy for each single value of the hyperparameter on the x-axis. The solid curves are the quadratic regression curves of prediction accuracy on the hyperparameter on the x-axis. The maximum prediction accuracy (dashed curves) is more important than the average accuracy (solid curves) because we target only top models rather than average models. The results for the validation set are available in Appendix IV. Overall, ASU-DNN could outperform F-DNN regardless of the values of the other hyperparameters.}
\label{fig:ah}
\end{figure*}

\noindent
We further examine whether the alternative-specific connectivity hyperparameter is more effective than the other hyperparameters, including explicit regularizations, implicit regularizations, and architectural hyperparameters. Figure \ref{fig:ah} shows the results, with each of the subfigures depicting the comparison of a hyperparameter with the alternative-specific connectivity hyperparameter.

\textbf{Explicit regularizations.} Figures \ref{sfig:l1_test} and \ref{sfig:l2_test} show how the prediction accuracy varies with the alternative-specific connectivity hyperparameter and two hyperparameters of explicit regularizations: $l_1$ and $l_2$ penalties. The $2-3 \%$ prediction accuracy gain by ASU-DNN is retained across the different values of the $l_1$ and $l_2$ regularizations. When the $l_1$ penalty is smaller than $10^{-5}$ and $l_2$ penalty is smaller than $10^{-3}$, ASU-DNN exhibits consistently higher prediction accuracy than F-DNN does. The $l_1$ and $l_2$ regularizations fail to aid in achieving a higher prediction accuracy by either ASU-DNN and F-DNN, as illustrated by the nearly flat maximum prediction accuracy curve when $l_1$ and $l_2$ values are small and a large decrease in the prediction accuracy as $l_1$ and $l_2$ increase, in both Figures \ref{sfig:l1_test} and \ref{sfig:l2_test}. In other words, the most commonly used $l_1$ and $l_2$ regularizations cannot aid model prediction, or at least they are less effective than the alternative-specific connectivity hyperparameter. 

\textbf{Implicit regularizations.} Figures \ref{sfig:lr_test}, \ref{sfig:ni_test}, \ref{sfig:n_mini_test}, and \ref{sfig:bn_test} show the relationship between the alternative-specific connectivity hyperparameter and four implicit regularizations: learning rates, number of total iterations, size of mini batch, and batch normalization. These regularization methods are implicit because they are not explicitly used in the empirical risk minimization in Equation \ref{eq:erm}, although they have impacts on model training through the computational process. Again, the prediction accuracy gain owing to the alternative-specific connectivity is highly robust regardless of the values of the other four hyperparameters: in all four figures, the dashed green curves are always placed higher than the dashed red curves are. In Figure \ref{sfig:lr_test}, both green and red curves assume a marginally concave quadratic form. The learning rates associated with the highest prediction accuracies are between $10^{-3}$ and $10^{-2}$, which are the default values in Tensorflow. This concave quadratic shape is intuitive because highly marginal learning rates are generally inadequate for achieving the optimum values and very large learning rates generally overshoot. In Figures \ref{sfig:ni_test}, \ref{sfig:n_mini_test}, and \ref{sfig:bn_test}, the dashed and solid curves of both F-DNN and ASU-DNN are nearly horizontal. This indicates that the number of iterations, size of mini batches, and batch normalization are immaterial for improving DNN's prediction accuracy in choice modeling tasks. 

\textbf{Architectural hyperparameters.} Figures \ref{sfig:M_test}, \ref{sfig:n_hidden_test}, and \ref{sfig:dr_test} compare the alternative-specific connectivity hyperparameter to three architectural hyperparameters: depth and width of DNN, and dropout rates. Similarly, the $2-3 \%$ prediction accuracy gain remains over approximately the whole range of the architectural hyperparameters. In Figure \ref{sfig:M_test}, the green dashed line is consistently higher than the red dashed line for the majority of the M values (from three to ten). However, this result is not exactly true when the depth of DNN is very small or very large. It is worthnoting that the model performance increases dramatically from one-layer to three-layer ASU-DNN. This indicates that the IIA constraint is less restrictive than the linear specification of each alternative's utility conditioning on the IIA constraint. In Figure \ref{sfig:n_hidden_test}, the maximum prediction accuracy of F-DNN form almost horizontal lines everywhere. Finally, in Figure \ref{sfig:dr_test}, whereas the prediction accuracy difference remains approximately $2-3 \%$ for most of the values of the dropout rate, this difference becomes approximately $10 \%$ when the dropout rate is larger than $0.1$. The prediction accuracy of ASU-DNN increases marginally as the dropout rates increase, whereas that of F-DNN decreases. These results imply that the alternative-specific connectivity exerts an interaction effect of activating architectural hyperparameters, in addition to its first order effects of $2-3 \%$ prediction gain.

\subsection{Interpretation of ASU-DNN: Combining IIA and DNN}
\begin{figure}[t!]
\centering
\subfloat[MNL]{\includegraphics[width=0.3\linewidth]{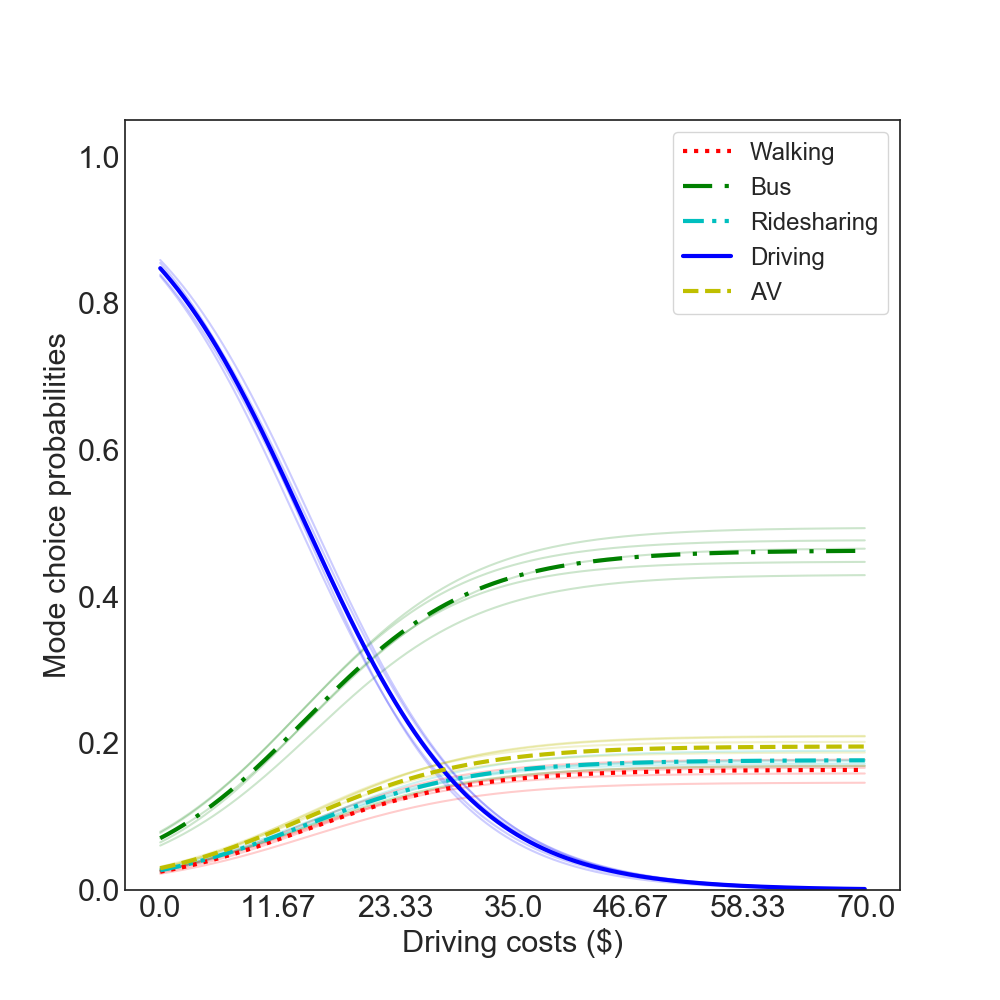}\label{sfig:interpretation_MNL_sgp}}
\subfloat[ASU-DNN (Top 1 Model)]{\includegraphics[width=0.3\linewidth]{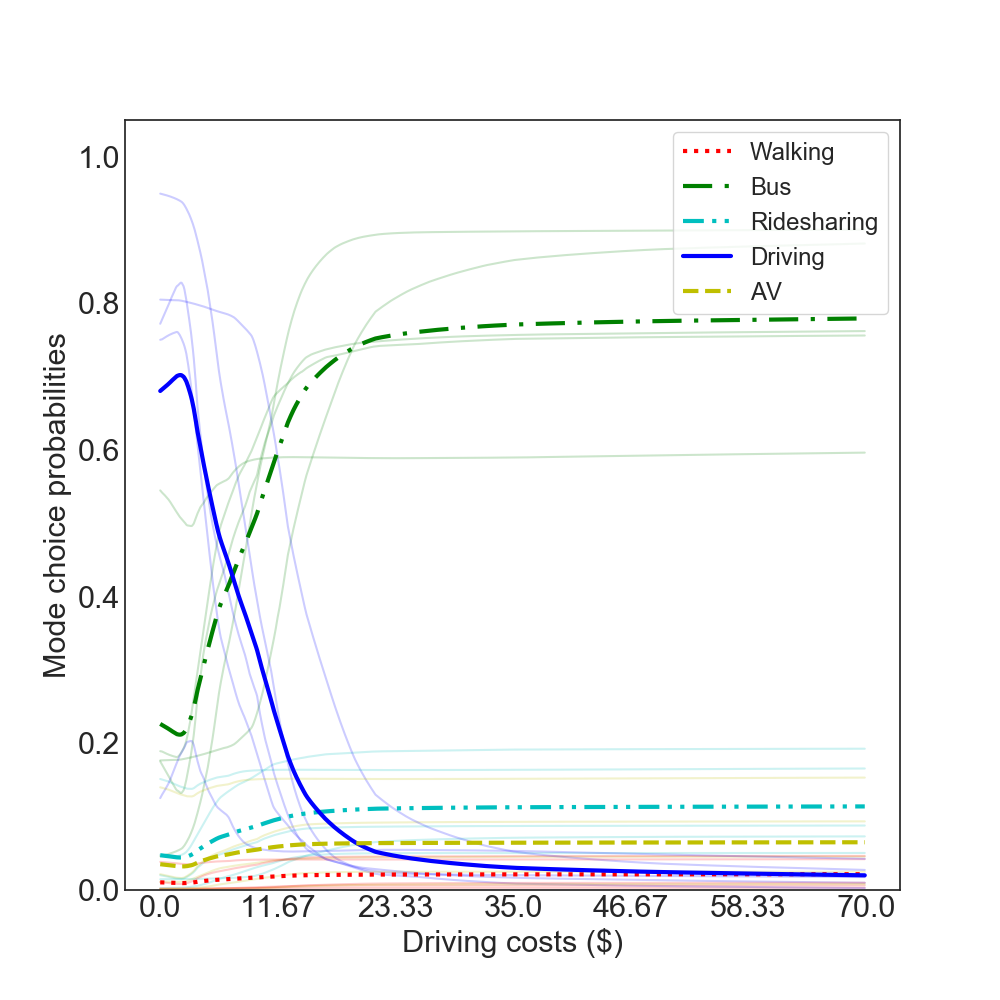}\label{sfig:interpretation_sparse_sgp_top1}}
\subfloat[ASU-DNN (Top 10 Models)]{\includegraphics[width=0.3\linewidth]{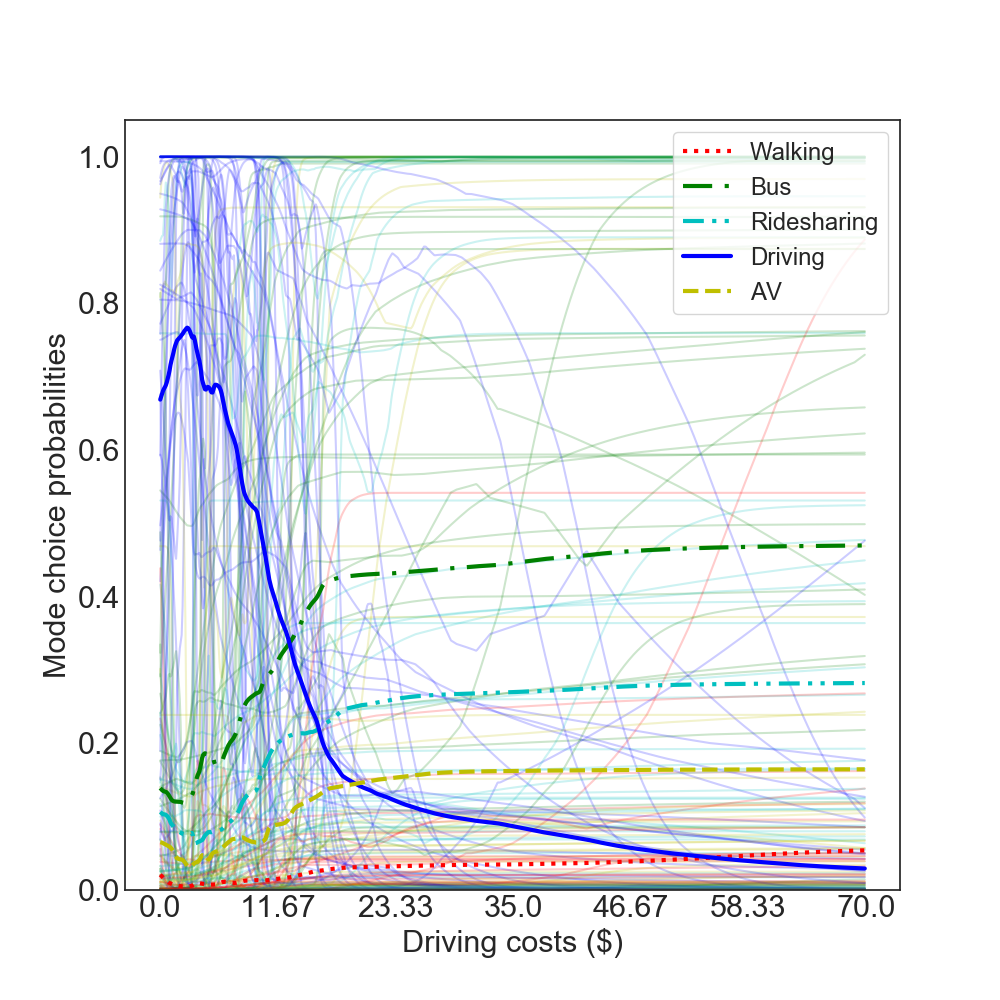}\label{fig:interpretation_sparse_sgp_top10}} \\
\subfloat[NL]{\includegraphics[width=0.3\linewidth]{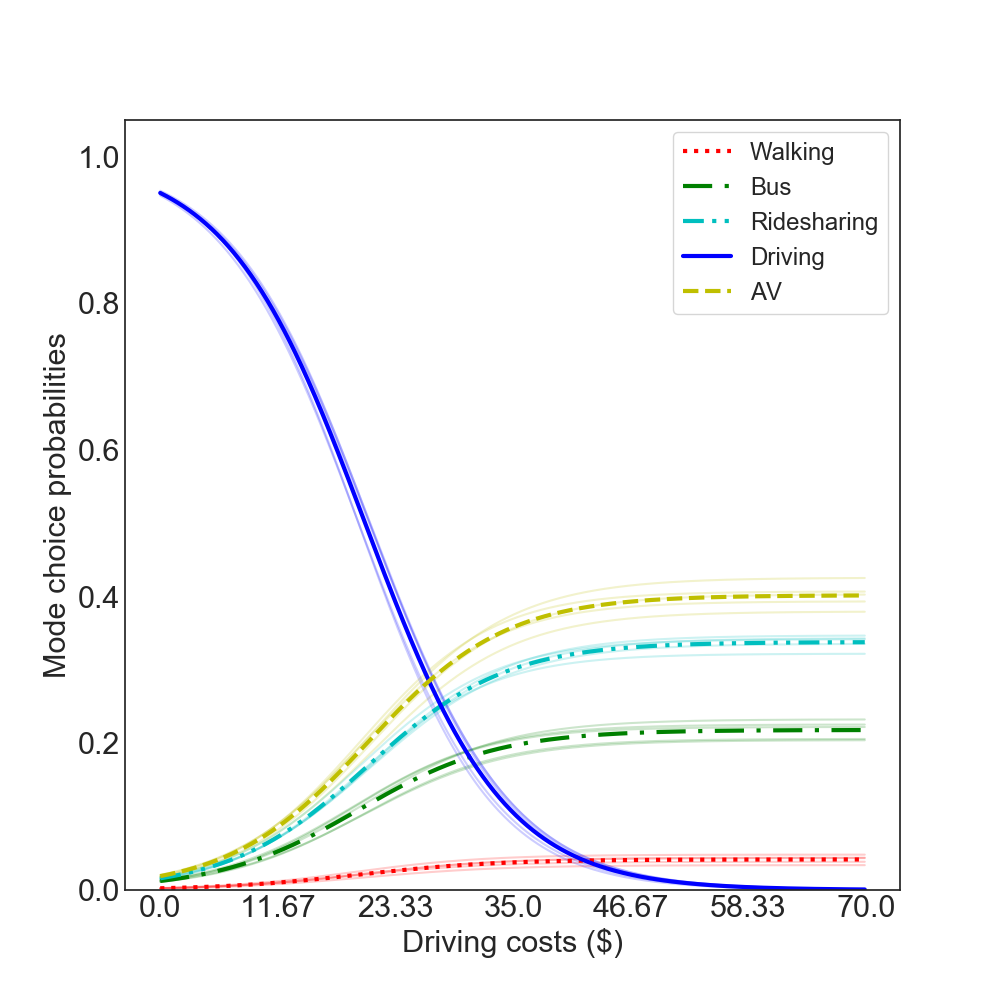}\label{sfig:interpretation_NL_sgp}} 
\subfloat[F-DNN (Top 1 Model)]{\includegraphics[width=0.3\linewidth]{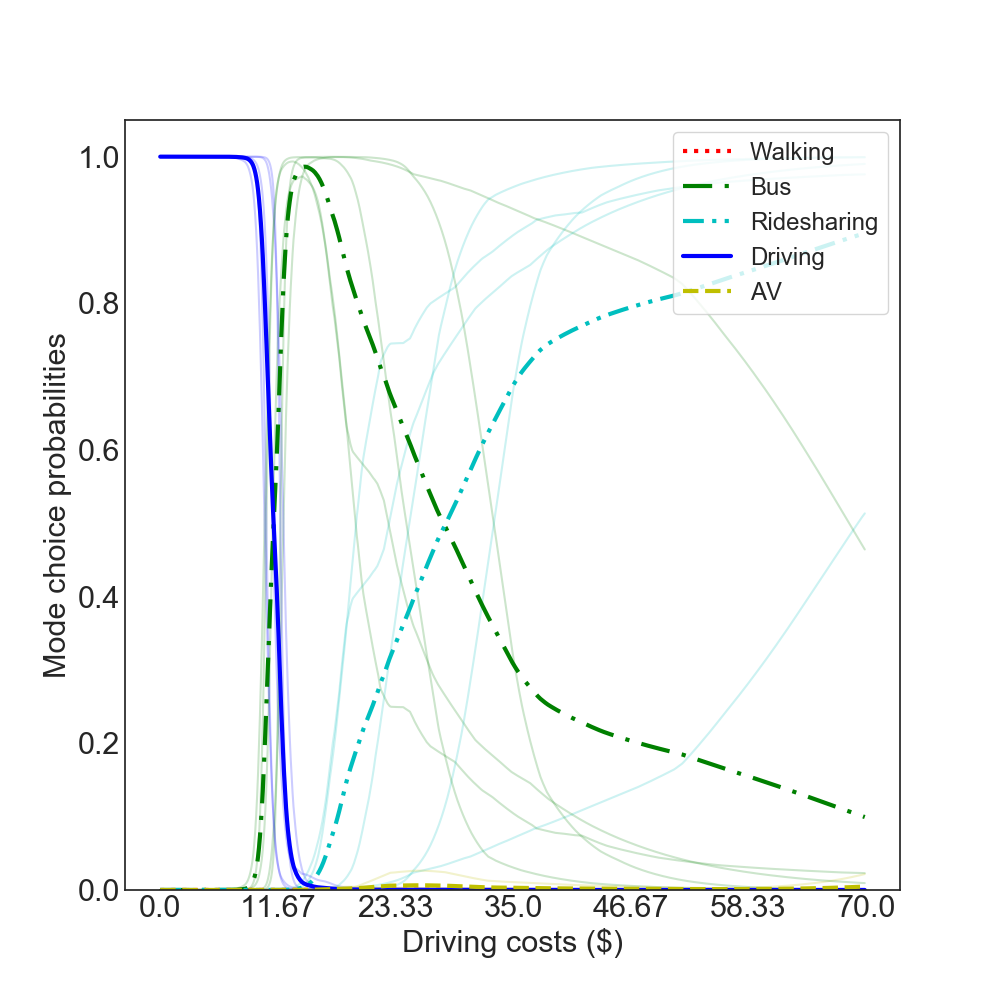}\label{sfig:interpretation_full_sgp_top1}} 
\subfloat[F-DNN (Top 10 Models)]{\includegraphics[width=0.3\linewidth]{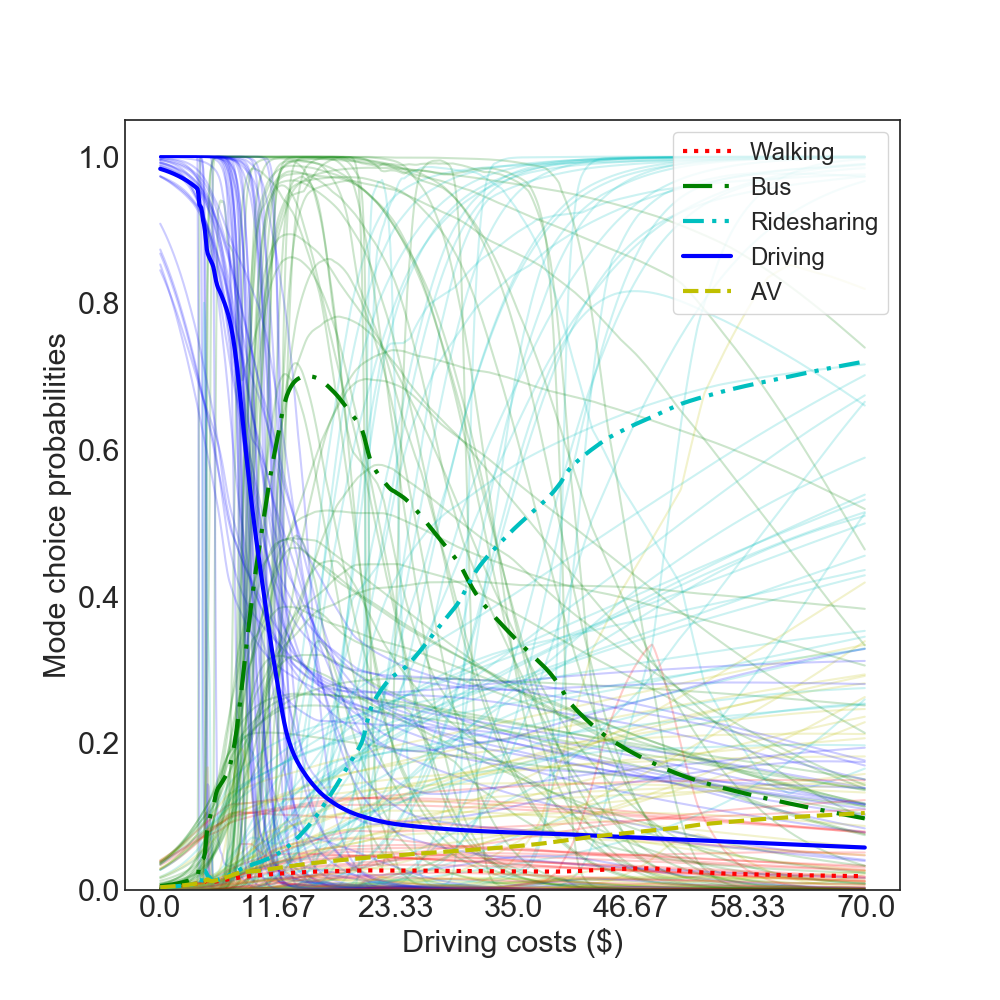}\label{sfig:interpretation_full_sgp_top10}} \\
\caption{Choice probability functions of MNL, NL, ASU-DNN, and F-DNN in the SGP testing set. \textit{Upper row:} MNL and ASU-DNN models. \textit{Lower row:} NL and F-DNN models. Each light curve represents a training result; the dark curves represent the average of all the training results. ASU-DNN compromises MNL and F-DNN, since it retains the global IIA-constraint substitution pattern of MNL and the local richness of F-DNN.}
\label{fig:interpretation}
\end{figure}

\noindent
Whereas DNN is generally criticized as lacking interpretability, we can visualize the choice probability functions and compute the elasticity coefficients in DNN models by using numerical simulation \cite{Bentz2000,Montavon2018,Baehrens2010,Ross2018,WangShenhao2018_ml2}. Figure \ref{fig:interpretation} shows how, following this method, the probabilities of selecting five travel modes vary with increasing driving costs in the ASU-DNN, F-DNN, MNL, and NL models, while holding all other variables constant at their empirical mean values. 

The choice probability functions of ASU-DNN mix the behavioral patterns of MNL and F-DNN, since ASU-DNN retains the global IIA-constraint substitution pattern from MNL and the local richness from F-DNN. Comparing ASU-DNN and F-DNN, the choice probability functions of ASU-DNN appear more intuitive than those of F-DNN for at least two reasons. The first difference is with regard to the substitution pattern between the five travel modes; specifically, F-DNN predicts that the probability of catching buses will decrease dramatically as the driving cost increases beyond $\$ 15$, whereas ASU-DNN predicts that this probability will increase marginally. The substitute effect between driving and catching buses predicted by ASU-DNN appears to be more reasonable, consistent with the common notion that the alternatives are often substitute goods. Note that the substitution pattern of travel modes in ASU-DNN describes that individuals could switch from driving to the other modes in a proportional manner, which is similar to the MNL model in Figure \ref{sfig:interpretation_MNL_sgp}. The second difference between ASU-DNN and F-DNN is in the probability of selecting driving as the driving costs approach zero. ASU-DNN predicts that individuals exhibit $70\%$ probability of selecting driving when driving costs become zero, whereas F-DNN predicts this probability being close to $100\%$. The latter value appears unreasonable because all the other variables including driving time is fixed as the mean value of the sample, resulting in the likelihood of the selection of alternative travel modes. Overall, ASU-DNN presents more regularity than F-DNN, which is caused by the built-in alternative-specific connectivity design. 

Tables \ref{table:MNL_elasticity}-\ref{table:F_DNN_elasticity} summarize the elasticity coefficients for the MNL, NL, top 10 ASU-DNN, and top 10 F-DNN models, with negative values being bolded to highlight the structure in each table. These elasticity coefficients are computed by simulation, with each input variable varied by 1\% holding all the other variables constant at the sample mean values. As shown in Table \ref{table:MNL_elasticity}, the MNL model clearly reveals its IIA substitution pattern in two ways. First, all the self-elasticity coefficients are negative as highlighted on the main diagonal, while the cross-elasticity coefficients are all positive. Second, the four cross-elasticity coefficients regarding one specific attribute have the same magnitude. For example, regarding the walking time, the cross-elasticity coefficients of taking buses, ride-sharing, driving, and using AVs are all $0.134$, which is consistent with the elasticity formula of MNL models \footnote{Please refer to Chapter 3 in Train's textbook \cite{Train2009}}. Table \ref{table:NL_elasticity} demonstrates how a NL model has a more flexible substitution pattern than MNL. The elasticity coefficients take a clear block-wise shape and the values within a nest are different from those cross nests.

\begin{table}[ht!]
\caption{Elasticity coefficients of MNL}
\centering
\resizebox{0.75\textwidth}{!}{
\begin{tabular}{l|lllll}
\hline
                             & Walk   & Bus    & Ridesharing & Drive  & AV     \\ \hline
Walk: walk time              & \textbf{-1.890} & 0.134  & 0.134       & 0.134  & 0.134  \\
Bus: cost                    & 0.137  & \textbf{-0.546} & 0.137       & 0.137  & 0.137  \\
Bus: in-vehicle time         & 0.128  & \textbf{-0.475} & 0.128       & 0.128  & 0.128  \\
Ridesharing: cost            & 0.029  & 0.029  & \textbf{-0.240}      & 0.029  & 0.029  \\
Ridesharing: in-vehicle time & 0.083  & 0.083  & \textbf{-0.740}      & 0.083  & 0.083  \\
Drive: cost                  & 0.288  & 0.288  & 0.288       & \textbf{-0.793} & 0.288  \\
Drive: in-vehicle time       & 0.280  & 0.280  & 0.280       & \textbf{-0.440} & 0.280  \\
AV: cost                     & 0.048  & 0.048  & 0.048       & 0.048  & \textbf{-0.449} \\
AV: in-vehicle time          & 0.060  & 0.060  & 0.060       & 0.060  & \textbf{-0.560} \\ \hline
\end{tabular}
}
\label{table:MNL_elasticity}
\end{table}

\begin{table}[htb]
\caption{Elasticity coefficients of NL. Nest 1: walk and bus. Nest 2: ridesharing, drive, and AV}
\centering
\resizebox{0.75\textwidth}{!}{
\begin{tabular}{l|lllll}
\hline
                             & Walk   & Bus    & Ridesharing & Drive  & AV     \\ \hline
Walk: walk time              & \textbf{-1.481} & \textbf{-0.067} & 0.163       & 0.163  & 0.163  \\
Bus: cost                    & \textbf{-0.074} & \textbf{-0.550} & 0.170       & 0.170  & 0.170  \\
Bus: in-vehicle time         & \textbf{-0.050} & \textbf{-0.356} & 0.116       & 0.116  & 0.116  \\
Ridesharing: cost            & 0.039  & 0.039  & \textbf{-0.488}      & 0.080  & 0.080  \\
Ridesharing: in-vehicle time & 0.073  & 0.073  & \textbf{-0.988}      & 0.146  & 0.146  \\
Drive: cost                  & 0.229  & 0.229  & 0.424       & \textbf{-0.905} & 0.424  \\
Drive: in-vehicle time       & 0.269  & 0.269  & 0.482       & \textbf{-0.593} & 0.482  \\
AV: cost                     & 0.048  & 0.048  & 0.093       & 0.093  & \textbf{-0.687} \\
AV: in-vehicle time          & 0.057  & 0.057  & 0.109       & 0.109  & \textbf{-0.800} \\ \hline
\end{tabular}
}
\label{table:NL_elasticity}
\end{table}

The elasticity coefficients in Table \ref{table:ASU_DNN_elasticity} shows that the substitution pattern of ASU-DNN is very similar to MNL in Table \ref{table:MNL_elasticity}. The similarity can be seen from the positive self-elasticity coefficients on the main diagonal, the negative cross-elasticity coefficients on the off-diagonal, and the same cross-elasticity coefficients regarding one specific attribute. This similarity should not be a surprise, since the ASU-DNN in the family of DNN models corresponds to the MNL in the family of discrete choice models, owing to the alternative-specific utility functions in ASU-DNN. As a comparison, the elasticity coefficients of F-DNN in Table \ref{table:F_DNN_elasticity} are much more irregular than those of ASU-DNN and even NL: many cross-elasticity coefficients are negative and the elasticity coefficients don't have the block-wise pattern as in NL. Note that this ``irregularity'' in F-DNN does not necessarily have a negative connotation. It can be the case that the elasticity pattern in F-DNN captures the real data generating process that is out of the model families of MNL, NL, or even ASU-DNN. Therefore, F-DNN might enable researchers to capture the highly correlated utility errors, as in mixed logit (MXL) models. However, it is hard to make a definitive judgment by using only our empirical results. We leave these two questions, whether F-DNN describes the highly correlated utility error terms (as in MXL) and whether the behavioral patterns revealed in ASU-DNN and F-DNN are realistic, open to future studies.

\begin{table}[htb]
\caption{Average elasticity coefficients of top 10 ASU-DNN Models}
\centering
\resizebox{0.75\textwidth}{!}{
\begin{tabular}{l|lllll}
\hline
                             & Walk    & Bus    & Ridesharing & Drive  & AV     \\ \hline
Walk: walk time              & \textbf{-10.016} & 1.029  & 1.028       & 1.029  & 1.030  \\
Bus: cost                    & 0.381   & \textbf{-1.983} & 0.395       & 0.396  & 0.391  \\
Bus: in-vehicle time         & 0.440   & \textbf{-3.198} & 0.438       & 0.435  & 0.436  \\
Ridesharing: cost            & 0.219   & 0.221  & \textbf{-2.638}      & 0.221  & 0.223  \\
Ridesharing: in-vehicle time & 0.420   & 0.421  & \textbf{-4.878}      & 0.420  & 0.420  \\
Drive: cost                  & 1.709   & 1.735  & 1.726       & \textbf{-2.249} & 1.731  \\
Drive: in-vehicle time       & 2.138   & 2.172  & 2.178       & \textbf{-1.952} & 2.171  \\
AV: cost                     & 0.383   & 0.379  & 0.380       & 0.380  & \textbf{-4.681} \\
AV: in-vehicle time          & 0.364   & 0.362  & 0.363       & 0.362  & \textbf{-3.485} \\ \hline
\end{tabular}
}
\label{table:ASU_DNN_elasticity}
\end{table}

\begin{table}[htb]
\caption{Average elasticity coefficients of top 10 F-DNN Models}
\centering
\resizebox{0.75\textwidth}{!}{
\begin{tabular}{l|lllll}
\hline
                             & Walk   & Bus    & Ridesharing & Drive  & AV     \\ \hline
Walk: walk time              & \textbf{-4.228} & 0.580  & 0.447       & 0.172  & 0.109  \\
Bus: cost                    & \textbf{-0.696} & \textbf{-2.052} & \textbf{-0.093}      & 0.623  & 0.342  \\
Bus: in-vehicle time         & \textbf{-0.053} & \textbf{-1.803} & \textbf{-0.339}      & 0.502  & 0.588  \\
Ridesharing: cost            & 0.055  & 0.292  & \textbf{-1.858}      & 0.142  & 1.457  \\
Ridesharing: in-vehicle time & \textbf{-0.139} & \textbf{-0.115} & \textbf{-3.436}      & 0.434  & 0.268  \\
Drive: cost                  & 0.897  & 1.404  & 2.079       & \textbf{-1.711} & 1.474  \\
Drive: in-vehicle time       & 1.266  & 1.690  & 2.164       & \textbf{-1.748} & 1.937  \\
AV: cost                     & \textbf{-0.516} & 0.036  & 0.356       & 0.443  & \textbf{-3.781} \\
AV: in-vehicle time          & \textbf{-0.769} & 0.457  & 0.074       & 0.360  & \textbf{-3.288} \\ \hline
\end{tabular}
}
\label{table:F_DNN_elasticity}
\end{table}

\section{Conclusion and Discussion}
\label{s:7}
\noindent
This study is motivated by the challenges in the application of DNN to choice analysis, including the tension between domain-specific knowledge and generic-purpose models, and the lack of interpretability and effective regularization methods. In contrast to most of the recent studies in the transportation domain that straightforwardly apply various DNN models to choice analysis, we demonstrate that the benefit could flow in the other direction: from domain knowledge to DNN models. Specifically, it is feasible to inject behavioral insights into DNN architecture owing to the implicit RUM interpretation in DNN. By using the alternative-specific utility constraint, we design a new DNN architecture ASU-DNN, which achieves a certain compromise between domain-specific knowledge and generic-purpose DNN, and between the handcrafted feature learning and automatic feature learning paradigms. This compromise is significantly effective, as demonstrated by our empirical results that ASU-DNN model is more predictive and provides more regular behavioral information than F-DNN. ASU-DNN could outperform F-DNN by approximately $2-3\%$ in both validation and testing data sets regardless of the values of DNN’s other hyperparameters. The behavioral insights from ASU-DNN are also more reasonable than those from F-DNN, as shown in the choice probability functions of the five travel modes. Theoretically, this alternative-specific utility specification leads to the IIA constraint, which can be considered as a regularization method under the DNN framework. This constraint causes the DNN architecture to be sparser, resulting in a lower estimation error. This insight is supported by our empirical result, because the alternative-specific utility constraint as a domain-knowledge-based regularization is more effective than other explicit and implicit regularization methods, and architectural hyperparameters. In addition, the comparison between ASU-DNN and F-DNN could function as a behavioral test, and our results indicate that individuals are more likely to compute the utility based on an alternative’s own attributes rather than the attributes of all the alternatives. This finding is consistent with the long-standing practice in choice modeling.

One natural question is to what extent our findings are generalizable. This ASU-DNN model is guaranteed to have the IIA-constraint substitution pattern, smaller estimation errors than F-DNN, and more flexibility and higher function approximation power than MNL. These results are always generaliable, owing to the design of ASU-DNN architecture. However, it is neither theoretically nor empirically guaranteed that ASU-DNN always outperforms F-DNN and MNL in terms of prediction accuracy. The prediction performance depends on the sample size, model complexity, and the underlying data generating process (DGP) that is never known to researchers in empirical studies. Loosely speaking, ASU-DNN tends to perform better than F-DNN when sample size becomes smaller, DGP is closer to the IIA-constraint substitution pattern, and the number of alternatives in the choice set becomes larger. ASU-DNN tends to outperform MNL when the utility specification of each alternative is more complicated than simple linear or quadratic forms, although both ASU-DNN and MNL will have misspecification errors when the true DGP deviates from the alternative-specific utility specification. Related to this generalizability discussion, another open question is whether the behavioral pattern revealed in ASU-DNN is realistic. Unfortunately this realism question is hard to answer given that the DGP is never known to researchers in empirical studies. Instead of making a value judgment here, we would encourage future studies to use simulations to answer under what conditions ASU-DNN can approximate the true DGP in a more efficient manner than both F-DNN and MNL.

The alternative-specific utility specification can be incorrect in ASU-DNN. However, it is important to note that this problem exists in any modeling practice because any prior knowledge could be incorrect. The method of using prior knowledge in ASU-DNN is fundamentally different from that in traditional choice models. ASU-DNN starts with a universal approximator F-DNN as a baseline and ``builds downward'' F-DNN by using only a piece of prior knowledge (alternative-specific utility in this study) to reduce the complexity of F-DNN. In contrast, traditional choice modeling starts from scratch as a baseline and ``builds upwards'' a choice model by using all types of prior knowledge (e.g. linearity and additivity of utilities). The former is a significantly more conservative method of using prior knowledge. As a result, the downward-built models are more robust to the function misspecification problem.

The ASU-DNN in the family of DNN models is the counterpart of the MNL in the family of discrete choice models. This mapping is enabled by a triangle relationship between the IIA-constraint substitution pattern, choice probability functions taking the Softmax form, and the IID error terms with extreme value distributions. This triangle relationship was neatly established in McFadden's seminal paper \cite{McFadden1974}, which demonstrates any one of the three conditions leads to the other two under the RUM framework. However, the triangle relationship does not explicitly exist for choice models beyond MNL. Whereas researchers can derive the choice probability functions of NL based on the generalized extreme value (GEV) distributions, the proof of the reversed direction is unclear. The mixed logit (MXL) model that allows more flexible correlation between the utility error terms is even more complicated, since the choice probabilities in MXL are computed by sampling, which deviates further away from any analytical approach. Our study has empirically demonstrated that the elasticity coefficients of F-DNN are more flexible than NL as shown in Tables \ref{table:NL_elasticity} and \ref{table:F_DNN_elasticity}, leading to our conjecture that there exists another regularized DNN model that is corresponding to the NL or GEV models. A valid support for this conjecture is beyond the scope of this study, and we hope future studies can identify the regularization methods that are associated with the NL or even MXL models.

Regardless of certain caveats and remaining questions, our results are promising because they present a solution to many challenges in DNN applications. More importantly, it indicates a new research direction of using utility theory to design DNN architectures for choice models, which could become more predictive owing to lower estimation errors and be more interpretable owing to the knowledge introduced into DNN as regularization. We consider that this research direction has immense potential because both utility theory and DNN architectures are exceptionally rich and active research fields. The alternative-specific utility connectivity is only a tiny piece among a vast number of insights in utility theory. Therefore, the immediate next steps could be to use more flexible utility functions (such as those in NL and MXL) to design novel DNN architectures. Future researchers should also examine the generalizability of ASU-DNN by testing whether it can perform better than F-DNN and choice models in other contexts.

\section*{Author Contributions}
\noindent
S.W. conceived of the presented idea, developed the theory, reviewed previous studies, and derived the analytical proofs. S.W. and B.M. designed and conducted the experiments; S.W. drafted the manuscripts; J.Z. provided comments and supervised this work. All authors discussed the results and contributed to the final manuscript.

\printbibliography

@article{ZhouBolei2014,
   Author = {Zhou, Bolei and Khosla, Aditya and Lapedriza, Agata and Oliva, Aude and Torralba, Antonio},
   Title = {Object detectors emerge in deep scene cnns},
   Journal = {arXiv preprint arXiv:1412.6856},
   Year = {2014} }

@article{ZhangChiyuan2016,
   Author = {Zhang, Chiyuan and Bengio, Samy and Hardt, Moritz and Recht, Benjamin and Vinyals, Oriol},
   Title = {Understanding deep learning requires rethinking generalization},
   Journal = {arXiv preprint arXiv:1611.03530},
   Year = {2016} }

@article{XieChi2003,
   Author = {Xie, Chi and Lu, Jinyang and Parkany, Emily},
   Title = {Work travel mode choice modeling with data mining: decision trees and neural networks},
   Journal = {Transportation Research Record: Journal of the Transportation Research Board},
   Number = {1854},
   Pages = {50-61},
   Year = {2003} }

@article{Weaver2009,
   Author = {Weaver, Ray and Frederick, Shane},
   Title = {Transaction disutility and the endowment effect},
   Journal = {NA-Advances in Consumer Research Volume 36},
   Year = {2009} }

@article{Vapnik1999,
   Author = {Vapnik, Vladimir Naumovich},
   Title = {An overview of statistical learning theory},
   Journal = {IEEE transactions on neural networks},
   Volume = {10},
   Number = {5},
   Pages = {988-999},
   Year = {1999} }

@book{Train2009,
   Author = {Train, Kenneth E},
   Title = {Discrete choice methods with simulation},
   Publisher = {Cambridge university press},
   Year = {2009} }

@inproceedings{Szegedy2015,
   Author = {Szegedy, Christian and Liu, Wei and Jia, Yangqing and Sermanet, Pierre and Reed, Scott and Anguelov, Dragomir and Erhan, Dumitru and Vanhoucke, Vincent and Rabinovich, Andrew},
   Title = {Going deeper with convolutions},
   Publisher = {Cvpr},
   Year = {2015} }

@article{Sekhar2016,
   Author = {Sekhar, Ch Ravi and Madhu, E},
   Title = {Mode Choice Analysis Using Random Forrest Decision Trees},
   Journal = {Transportation Research Procedia},
   Volume = {17},
   Pages = {644-652},
   Year = {2016} }

@inproceedings{Ribeiro2016,
   Author = {Ribeiro, Marco Tulio and Singh, Sameer and Guestrin, Carlos},
   Title = {Why should i trust you?: Explaining the predictions of any classifier},
   BookTitle = {Proceedings of the 22nd ACM SIGKDD International Conference on Knowledge Discovery and Data Mining},
   Publisher = {ACM},
   Pages = {1135-1144},
   Year = {2016} }

@article{Pulugurta2013,
   Author = {Pulugurta, Sarada and Arun, Ashutosh and Errampalli, Madhu},
   Title = {Use of artificial intelligence for mode choice analysis and comparison with traditional multinomial logit model},
   Journal = {Procedia-Social and Behavioral Sciences},
   Volume = {104},
   Pages = {583-592},
   Year = {2013} }

@inproceedings{Paredes2017,
   Author = {Paredes, Miguel and Hemberg, Erik and O'Reilly, Una-May and Zegras, Chris},
   Title = {Machine learning or discrete choice models for car ownership demand estimation and prediction?},
   BookTitle = {Models and Technologies for Intelligent Transportation Systems (MT-ITS), 2017 5th IEEE International Conference on},
   Publisher = {IEEE},
   Pages = {780-785},
   Year = {2017} }

@article{Omrani2015,
   Author = {Omrani, Hichem},
   Title = {Predicting travel mode of individuals by machine learning},
   Journal = {Transportation Research Procedia},
   Volume = {10},
   Pages = {840-849},
   Year = {2015} }

@article{McFadden1974,
   Author = {McFadden, Daniel},
   Title = {Conditional logit analysis of qualitative choice behavior},
   Year = {1974} }

@article{Lipton2016,
   Author = {Lipton, Zachary C},
   Title = {The mythos of model interpretability},
   Journal = {arXiv preprint arXiv:1606.03490},
   Year = {2016} }

@article{LeCun2015,
   Author = {LeCun, Yann and Bengio, Yoshua and Hinton, Geoffrey},
   Title = {Deep learning},
   Journal = {Nature},
   Volume = {521},
   Number = {7553},
   Pages = {436-444},
   Year = {2015} }

@inproceedings{Krizhevsky2012,
   Author = {Krizhevsky, Alex and Sutskever, Ilya and Hinton, Geoffrey E},
   Title = {Imagenet classification with deep convolutional neural networks},
   BookTitle = {Advances in neural information processing systems},
   Pages = {1097-1105},
   Year = {2012} }

@article{Koh2017,
   Author = {Koh, Pang Wei and Liang, Percy},
   Title = {Understanding black-box predictions via influence functions},
   Journal = {arXiv preprint arXiv:1703.04730},
   Year = {2017} }

@article{Karlaftis2011,
   Author = {Karlaftis, Matthew G and Vlahogianni, Eleni I},
   Title = {Statistical methods versus neural networks in transportation research: Differences, similarities and some insights},
   Journal = {Transportation Research Part C: Emerging Technologies},
   Volume = {19},
   Number = {3},
   Pages = {387-399},
   Year = {2011} }

@article{Hornik1989,
   Author = {Hornik, Kurt and Stinchcombe, Maxwell and White, Halbert},
   Title = {Multilayer feedforward networks are universal approximators},
   Journal = {Neural networks},
   Volume = {2},
   Number = {5},
   Pages = {359-366},
   Year = {1989} }

@article{Hinton2012,
   Author = {Hinton, Geoffrey E and Srivastava, Nitish and Krizhevsky, Alex and Sutskever, Ilya and Salakhutdinov, Ruslan R},
   Title = {Improving neural networks by preventing co-adaptation of feature detectors},
   Journal = {arXiv preprint arXiv:1207.0580},
   Year = {2012} }

@inproceedings{HeKaiming2015_2,
   Author = {He, Kaiming and Zhang, Xiangyu and Ren, Shaoqing and Sun, Jian},
   Title = {Delving deep into rectifiers: Surpassing human-level performance on imagenet classification},
   BookTitle = {Proceedings of the IEEE international conference on computer vision},
   Pages = {1026-1034},
   Year = {2015} }

@article{Hagenauer2017,
   Author = {Hagenauer, Julian and Helbich, Marco},
   Title = {A comparative study of machine learning classifiers for modeling travel mode choice},
   Journal = {Expert Systems with Applications},
   Volume = {78},
   Pages = {273-282},
   Year = {2017} }

@book{Goodfellow2016,
   Author = {Goodfellow, Ian and Bengio, Yoshua and Courville, Aaron and Bengio, Yoshua},
   Title = {Deep learning},
   Publisher = {MIT press Cambridge},
   Volume = {1},
   Year = {2016} }

@inproceedings{Glorot2011,
   Author = {Glorot, Xavier and Bordes, Antoine and Bengio, Yoshua},
   Title = {Domain adaptation for large-scale sentiment classification: A deep learning approach},
   BookTitle = {Proceedings of the 28th international conference on machine learning (ICML-11)},
   Pages = {513-520},
   Year = {2011} }

@inproceedings{Glorot2010,
   Author = {Glorot, Xavier and Bengio, Yoshua},
   Title = {Understanding the difficulty of training deep feedforward neural networks},
   BookTitle = {Proceedings of the thirteenth international conference on artificial intelligence and statistics},
   Pages = {249-256},
   Year = {2010} }

@article{Freitas2014,
   Author = {Freitas, Alex A},
   Title = {Comprehensible classification models: a position paper},
   Journal = {ACM SIGKDD explorations newsletter},
   Volume = {15},
   Number = {1},
   Pages = {1-10},
   Year = {2014} }

@book{Dhami2016,
   Author = {Dhami, Sanjit},
   Title = {The Foundations of Behavioral Economic Analysis},
   Publisher = {Oxford University Press},
   Year = {2016} }

@article{Brauneis2017,
   Author = {Brauneis, Robert and Goodman, Ellen P},
   Title = {Algorithmic transparency for the smart city},
   Year = {2017} }

@article{Boshi_Velez2017,
   Author = {Doshi-Velez, Finale and Kim, Been},
   Title = {Towards a rigorous science of interpretable machine learning},
   Year = {2017} }

@article{Bengio2013,
   Author = {Bengio, Yoshua and Courville, Aaron and Vincent, Pascal},
   Title = {Representation learning: A review and new perspectives},
   Journal = {IEEE transactions on pattern analysis and machine intelligence},
   Volume = {35},
   Number = {8},
   Pages = {1798-1828},
   Year = {2013} }

@book{Ben_Akiva1985,
   Author = {Ben-Akiva, Moshe E and Lerman, Steven R},
   Title = {Discrete choice analysis: theory and application to travel demand},
   Publisher = {MIT press},
   Volume = {9},
   Year = {1985} }

@article{Bergstra2012,
   Author = {Bergstra, James and Bengio, Yoshua},
   Title = {Random search for hyper-parameter optimization},
   Journal = {Journal of Machine Learning Research},
   Volume = {13},
   Number = {Feb},
   Pages = {281-305},
      Year = {2012} }

@inproceedings{Joz2015,
   Author = {Jozefowicz, Rafal and Zaremba, Wojciech and Sutskever, Ilya},
   Title = {An empirical exploration of recurrent network architectures},
   BookTitle = {International Conference on Machine Learning},
   Pages = {2342-2350},
      Year = {2015} }

@article{Zoph2016,
   Author = {Zoph, Barret and Le, Quoc V},
   Title = {Neural architecture search with reinforcement learning},
   Journal = {arXiv preprint arXiv:1611.01578},
      Year = {2016} }

@article{Falkner2018,
   Author = {Falkner, Stefan and Klein, Aaron and Hutter, Frank},
   Title = {BOHB: Robust and efficient hyperparameter optimization at scale},
   Journal = {arXiv preprint arXiv:1807.01774},
      Year = {2018} }

@inproceedings{Snoek2015,
   Author = {Snoek, Jasper and Rippel, Oren and Swersky, Kevin and Kiros, Ryan and Satish, Nadathur and Sundaram, Narayanan and Patwary, Mostofa and Prabhat, Mr and Adams, Ryan},
   Title = {Scalable bayesian optimization using deep neural networks},
   BookTitle = {International Conference on Machine Learning},
   Pages = {2171-2180},
      Year = {2015} }

@inproceedings{HeKaiming2016, 
   Author = {He, Kaiming and Zhang, Xiangyu and Ren, Shaoqing and Sun, Jian},
   Title = {Deep residual learning for image recognition},
   BookTitle = {Proceedings of the IEEE conference on computer vision and pattern recognition},
   Pages = {770-778},
      Year = {2016} }

@article{WangShenhao2018_ml2,
   Author = {Wang, Shenhao and Zhao, Jinhua},
   Title = {Using Deep Neural Network to Analyze Travel Mode Choice With Interpretable Economic Information: An Empirical Example},
   Journal = {arXiv preprint arXiv:1812.04528},
      Year = {2018} }

@article{Cantarella2005,
   Author = {Cantarella, Giulio Erberto and de Luca, Stefano},
   Title = {Multilayer feedforward networks for transportation mode choice analysis: An analysis and a comparison with random utility models},
   Journal = {Transportation Research Part C: Emerging Technologies},
   Volume = {13},
   Number = {2},
   Pages = {121-155},
      Year = {2005} }

@article{LiuLijuan2017,
   Author = {Liu, Lijuan and Chen, Rung-Ching},
   Title = {A novel passenger flow prediction model using deep learning methods},
   Journal = {Transportation Research Part C: Emerging Technologies},
   Volume = {84},
   Pages = {74-91},
      Year = {2017} }

@article{Polson2017,
   Author = {Polson, Nicholas G and Sokolov, Vadim O},
   Title = {Deep learning for short-term traffic flow prediction},
   Journal = {Transportation Research Part C: Emerging Technologies},
   Volume = {79},
   Pages = {1-17},
      Year = {2017} }

@article{ZhangZhenhua2018,
   Author = {Zhang, Zhenhua and He, Qing and Gao, Jing and Ni, Ming},
   Title = {A deep learning approach for detecting traffic accidents from social media data},
   Journal = {Transportation research part C: emerging technologies},
   Volume = {86},
   Pages = {580-596},
      Year = {2018} }

@article{Cranenburgh2019,
   Author = {van Cranenburgh, Sander and Alwosheel, Ahmad},
   Title = {An artificial neural network based approach to investigate travellers’ decision rules},
   Journal = {Transportation Research Part C: Emerging Technologies},
   Volume = {98},
   Pages = {152-166},
      Year = {2019} }

@techreport{Qianli2018,
   Author = {Liao, Qianli and Poggio, Tomaso},
   Title = {When Is Handcrafting Not a Curse?},
      Year = {2018} }

@article{Mhaskar2016,
   Author = {Mhaskar, Hrushikesh and Liao, Qianli and Poggio, Tomaso},
   Title = {Learning functions: when is deep better than shallow},
   Journal = {arXiv preprint arXiv:1603.00988},
      Year = {2016} }

@article{Sham1995,
   Author = {Sham, PC and Curtis, D},
   Title = {An extended transmission/disequilibrium test (TDT) for multi‐allele marker loci},
   Journal = {Annals of human genetics},
   Volume = {59},
   Number = {3},
   Pages = {323-336},
      Year = {1995} }

@article{Borsch1988,
   Author = {Börsch-Supan, Axel and Pitkin, John},
   Title = {On discrete choice models of housing demand},
   Journal = {Journal of Urban Economics},
   Volume = {24},
   Number = {2},
   Pages = {153-172},
      Year = {1988} }

@article{Guadagni1983,
   Author = {Guadagni, Peter M and Little, John DC},
   Title = {A logit model of brand choice calibrated on scanner data},
   Journal = {Marketing science},
   Volume = {2},
   Number = {3},
   Pages = {203-238},
      Year = {1983} }

@article{Bentz2000,
   Author = {Bentz, Yves and Merunka, Dwight},
   Title = {Neural networks and the multinomial logit for brand choice modelling: a hybrid approach},
   Journal = {Journal of Forecasting},
   Volume = {19},
   Number = {3},
   Pages = {177-200},
      Year = {2000} }

@article{Montavon2018, 
   Author = {Montavon, Gregoire and Samek, Wojciech and Muller, Klaus-Robert},
   Title = {Methods for interpreting and understanding deep neural networks},
   Journal = {Digital Signal Processing},
   Volume = {73},
   Pages = {1-15},
      Year = {2018}}

@book{Anthony2009,
   Author = {Anthony, Martin and Bartlett, Peter L},
   Title = {Neural network learning: Theoretical foundations},
   Publisher = {cambridge university press},
      Year = {2009}}

@book{Vershynin2018,
   Author = {Vershynin, Roman},
   Title = {High-dimensional probability: An introduction with applications in data science},
   Publisher = {Cambridge University Press},
   Volume = {47},
      Year = {2018}}

@book{Wainwright2019,
   Author = {Wainwright, Martin J},
   Title = {High-dimensional statistics: A non-asymptotic viewpoint},
   Publisher = {Cambridge University Press},
   Volume = {48},
      Year = {2019}}

@article{Bartlett2002,
   Author = {Bartlett, Peter L and Mendelson, Shahar},
   Title = {Rademacher and Gaussian complexities: Risk bounds and structural results},
   Journal = {Journal of Machine Learning Research},
   Volume = {3},
   Number = {Nov},
   Pages = {463-482},
      Year = {2002}}

@article{Bartlett2017,
   Author = {Bartlett, Peter L and Harvey, Nick and Liaw, Chris and Mehrabian, Abbas},
   Title = {Nearly-tight VC-dimension and pseudodimension bounds for piecewise linear neural networks},
   Journal = {arXiv preprint arXiv:1703.02930},
      Year = {2017} }

@article{Golowich2017,
   Author = {Golowich, Noah and Rakhlin, Alexander and Shamir, Ohad},
   Title = {Size-independent sample complexity of neural networks},
   Journal = {arXiv preprint arXiv:1712.06541},
      Year = {2017} }

@inproceedings{Neyshabur2015,
   Author = {Neyshabur, Behnam and Tomioka, Ryota and Srebro, Nathan},
   Title = {Norm-based capacity control in neural networks},
   BookTitle = {Conference on Learning Theory},
   Pages = {1376-1401},
      Year = {2015} }

@article{Cybenko1989,
   Author = {Cybenko, George},
   Title = {Approximation by superpositions of a sigmoidal function},
   Journal = {Mathematics of control, signals and systems},
   Volume = {2},
   Number = {4},
   Pages = {303-314},
      Year = {1989} }

@article{Hornik1991,
   Author = {Hornik, Kurt},
   Title = {Approximation capabilities of multilayer feedforward networks},
   Journal = {Neural networks},
   Volume = {4},
   Number = {2},
   Pages = {251-257},
      Year = {1991}}

@article{Poggio2017, 
   Author = {Poggio, Tomaso and Mhaskar, Hrushikesh and Rosasco, Lorenzo and Miranda, Brando and Liao, Qianli},
   Title = {Why and when can deep-but not shallow-networks avoid the curse of dimensionality: a review},
   Journal = {International Journal of Automation and Computing},
   Volume = {14},
   Number = {5},
   Pages = {503-519},
      Year = {2017} }

@article{Baehrens2010, 
   Author = {Baehrens, David and Schroeter, Timon and Harmeling, Stefan and Kawanabe, Motoaki and Hansen, Katja and MÃžller, Klaus-Robert},
   Title = {How to explain individual classification decisions},
   Journal = {Journal of Machine Learning Research},
   Volume = {11},
   Number = {Jun},
   Pages = {1803-1831},
      Year = {2010} }

@inproceedings{Ross2018,
   Author = {Ross, Andrew Slavin and Doshi-Velez, Finale},
   Title = {Improving the adversarial robustness and interpretability of deep neural networks by regularizing their input gradients},
   BookTitle = {Thirty-second AAAI conference on artificial intelligence},
      Year = {2018}}

@article{Nijkamp1996,
   Author = {Nijkamp, Peter and Reggiani, Aura and Tritapepe, Tommaso},
   Title = {Modelling inter-urban transport flows in Italy: A comparison between neural network analysis and logit analysis},
   Journal = {Transportation Research Part C: Emerging Technologies},
   Volume = {4},
   Number = {6},
   Pages = {323-338},
      Year = {1996} }

@article{Rao1998,
   Author = {Rao, PV Subba and Sikdar, PK and Rao, KV Krishna and Dhingra, SL},
   Title = {Another insight into artificial neural networks through behavioural analysis of access mode choice},
   Journal = {Computers, environment and urban systems},
   Volume = {22},
   Number = {5},
   Pages = {485-496},
      Year = {1998} }

@article{WuXin2018,
   author = {Wu, Xin and Guo, Jifu and Xian, Kai and Zhou, Xuesong},
   title = {Hierarchical travel demand estimation using multiple data sources: A forward and backward propagation algorithmic framework on a layered computational graph},
   journal = {Transportation Research Part C: Emerging Technologies},
   volume = {96},
   pages = {321-346},
   ISSN = {0968-090X},
   year = {2018},
   type = {Journal Article}
}

@article{SunJianping2019,
   author = {Sun, Jianping and Guo, Jifu and Wu, Xin and Zhu, Qian and Wu, Danting and Xian, Kai and Zhou, Xuesong},
   title = {Analyzing the Impact of Traffic Congestion Mitigation: From an Explainable Neural Network Learning Framework to Marginal Effect Analyses},
   journal = {Sensors},
   volume = {19},
   number = {10},
   pages = {2254},
   year = {2019},
   type = {Journal Article}
}

\newpage
\section*{Appendix I. Proof of Propositions 1 and 2}
\label{appendix:proof_mcfadden}
\noindent
\textbf{Proof of Proposition \ref{prop:1}.} This proof can be found in all choice modeling textbooks \cite{Train2009,Ben_Akiva1985}. With Gumbel distributional assumption, Equation \ref{eq:prob_1} could be solved in an analytical way:

\begin{equation}
\setlength{\jot}{2pt} \label{eq:choice_prob_deriv}
  \begin{aligned}
  P_{ik} &= \int_{- \infty}^{+\infty} \underset{j \neq k}{\prod} e^{-e^{-(V_{ik} - V_{ij} + \epsilon_{ik})}} f(\epsilon_{ik})d\epsilon_{ik} \\
         &= \int \underset{j}{\prod} e^{-e^{-(V_{ik} - V_{ij} + \epsilon_{ik})}} e^{- \epsilon_{ik}} d\epsilon_{ik} \\
         &= \int exp(e^{- \epsilon_{ik}} \underset{j}{\sum} {-e^{-(V_{ik} - V_{ij})}}) e^{- \epsilon_{ik}} d\epsilon_{ik} \\
         &= \int_{\infty}^{0} exp(-t \underset{j}{\sum} e^{-(V_{ik} - V_{ij})} ) dt \\
         &= \frac{e^{V_{ik}}}{\underset{j}{\sum} e^{V_{ij}}}
  \end{aligned}
\end{equation}

\noindent
in which the fourth equation uses $t = e^{- \epsilon_{ik}}$. Note this formula in Equation \ref{eq:choice_prob_deriv} is the Softmax function in DNN. $V_{ik}$ is both the deterministic utility in RUM and the inputs into the Softmax function in DNN. 

\noindent
\textbf{Proof of Proposition \ref{prop:2}.} This proof can be found in lemma 2 of Mcfadden (1974) \cite{McFadden1974}. Here is a brief summary of the proof. Suppose that one individual $i$ firstly chooses between alternative $k$ and $T$ alternatives $j$. Then according to Equations \ref{eq:prob_1} and \ref{eq:choice_prob_deriv}, 

\begin{equation}
\setlength{\jot}{2pt} \label{eq:choice_ik_1}
  \begin{aligned}
  P_{ik} &= \frac{e^{V_{ik}}}{e^{V_{ik}} + T e^{V_{ij}}} \\
         &= \int F(\epsilon_{ik} + V_{ik} - V_{ij})^T d F(\epsilon_{ik})
  \end{aligned}
\end{equation}

\noindent
Suppose that the individual $i$ chooses between alternatives $k$ and alternative $l$ in another choice scenario, and alternative $l$ is constructed such that $T e^{V_{ij}} = e^{V_{il}}$. Then

\begin{equation}
\setlength{\jot}{2pt} \label{eq:choice_ik_2}
  \begin{aligned}
  P_{ik} &= \frac{e^{V_{ik}}}{e^{V_{ik}} + e^{V_{il}}} \\
         &= \int F(\epsilon_{ik} + V_{ik} - V_{il}) d F(\epsilon_{ik}) \\
         &= \int F(\epsilon_{ik} + V_{ik} - V_{ij} - log T) d F(\epsilon_{ik}) 
  \end{aligned}
\end{equation}

\noindent
By construction, Equations \ref{eq:choice_ik_1} and \ref{eq:choice_ik_2} are equivalent
$$ \int F(\epsilon_{ik} + V_{ik} - V_{ij} - log T) - F(\epsilon_{ik} + V_{ik} - V_{ij})^T d F(\epsilon_{ik})  = 0 $$

\noindent
Since $F(\epsilon)$ is transition complete, meaning that $\forall a$, $E h(\epsilon + a) = 0$ implies $h(\epsilon) = 0, \forall \epsilon$, it implies 

$$F(V_{ik} - log \ T) = F(V_{ik})^T, \forall V_{ik}, T$$

\noindent
Taking $V_{ik} = 0$ implies $F(-log \ T) = e^{-\alpha T}$. Taking $V_{ik} = log \ T - log \ L$ implies $F(-log \ L) = F(log \ T/L)^T$. Hence $F(log \ T/L) = F(-log \ L)^{1/T} = e^{- \alpha L/T}$. Therefore, $F(\epsilon) = e^{-\alpha e^{-\epsilon}}$. This is the function of Gumbel distribution when $\alpha = 1$.

\section*{Appendix II. Summary Statistics of SGP and TRAIN}

\label{appendix:summary_stat}
\begin{table}[H]
\caption{Summary Statistics of SGP data set}
\centering
\resizebox{\textwidth}{!}{
\begin{tabular}{llllll}
\hline
\multicolumn{3}{l|}{\textbf{Variables}}                                                                & \multicolumn{3}{l}{\textbf{Variables}}                                                               \\ \hline
Name                            & Mean                   & \multicolumn{1}{l|}{Std.}                   & Name                                                & Mean                   & Std.                  \\\hline
Male (Yes = 1)                  & 0.383                  & \multicolumn{1}{l|}{0.486}                  & Age \textless 35 (Yes = 1)                          & 0.329                  & 0.470                 \\
Age\textgreater{}60 (Yes = 1)   & 0.075                  & \multicolumn{1}{l|}{0.263}                  & Low education (Yes = 1)                             & 0.331                  & 0.471                 \\
High education (Yes = 1)        & 0.480                  & \multicolumn{1}{l|}{0.500}                  & Low income (Yes = 1)                                & 0.035                  & 0.184                 \\
High income (Yes = 1)           & 0.606                  & \multicolumn{1}{l|}{0.489}                  & Full job (Yes = 1)                                  & 0.602                  & 0.490                 \\
Walk: walk time (min)           & 60.50                  & \multicolumn{1}{l|}{54.88}                  & Bus: cost (\$SG)                                    & 2.070                  & 1.266                 \\
Bus: walk time (min)            & 11.96                  & \multicolumn{1}{l|}{10.78}                  & Bus: waiting time (min)                             & 7.732                  & 5.033                 \\
Bus: in-vehilce time (min)      & 25.06                  & \multicolumn{1}{l|}{18.91}                  & RideSharing: cost (\$SG)                            & 14.48                  & 11.64                 \\
RideSharing: waiting time (min) & 7.108                  & \multicolumn{1}{l|}{4.803}                  & RideSharing: in-vehilce time (min)                  & 18.28                  & 13.39                 \\
AV: cost (\$SG)                 & 16.08                  & \multicolumn{1}{l|}{14.60}                  & AV: waiting time (min)                              & 7.249                  & 5.674                 \\
AV: in-vehilce time (min)       & 20.11                  & \multicolumn{1}{l|}{16.99}                  & Drive: cost (\$SG)                                  & 10.49                  & 10.57                 \\
Drive: walk time (min)          & 3.968                  & \multicolumn{1}{l|}{4.176}                  & Drive: in-vehilce time (min)                        & 17.43                  & 14.10                 \\ \hline
\multicolumn{6}{l}{\textbf{Statitics}}                                                                                                                                                                        \\ \hline
Number of samples               & \multicolumn{5}{l}{8418}                                                                                                                                                    \\
Number of choices               & \multicolumn{5}{l}{\begin{tabular}[c]{@{}l@{}}Walk: 874 (10.38\%); Bus: 1951 (23.18\%); RideSharing: 904 (10.74\%);\\ Drive 3774 (44.83\%); AV: 915 (10.87\%)\end{tabular}} \\ \hline
\end{tabular}}
\end{table}

\begin{table}[H]
\caption{Summary Statistics of TRAIN data set}
\centering
\resizebox{\textwidth}{!}{
\begin{tabular}{llllll} 
\hline
\multicolumn{6}{l}{\textbf{Variables}}                                                                                         \\ \hline
Name                              & Mean   & \multicolumn{1}{l|}{Std.}   & Name                              & Mean   & Std.   \\ \hline
Choice1: price (guilders)         & 3368.3 & \multicolumn{1}{l|}{1296.6} & Choice2: price (guilders)         & 3367.7 & 1274.3 \\
Choice1: time (min)               & 127.52 & \multicolumn{1}{l|}{29.13}  & Choice2: time (min)               & 127.17 & 27.96  \\
Choice1: number of changes        & 0.664  & \multicolumn{1}{l|}{0.733}  & Choice2: number of changes        & 0.681  & 0.743  \\
Choice1: comfort level (0,1 or 2) & 0.899  & \multicolumn{1}{l|}{0.602}  & Choice2: comfort level (0,1 or 2) & 0.885  & 0.617  \\ \hline
\multicolumn{6}{l}{\textbf{Statistics}}                                                                                        \\ \hline
Number of samples                 & \multicolumn{5}{l}{2928}                                                                   \\
Number of choices                 & \multicolumn{5}{l}{Choice1: 1473 (50.31\%); Choice2: 1455 (49.69\%)}                       \\ \hline
\end{tabular}
}
\end{table}

\section*{Appendix III. Top Five DNN Architectures}
\label{appendix:top_five_arch}
\begin{table}[H]
\caption{Top 5 DNN structures in the SGP data set}
\centering
\resizebox{\textwidth}{!}{
\begin{tabular}{l|lllll|lllll}
\hline
                           & \multicolumn{5}{c|}{F-DNN}                                   & \multicolumn{5}{c}{ASU-DNN}                                   \\ \hline
Rank                       & 1          & 2          & 3         & 4         & 5          & 1          & 2          & 3         & 4          & 5          \\
Accuracy (validation)      & 0.615      & 0.612      & 0.609     & 0.608     & 0.607      & 0.651      & 0.636      & 0.634     & 0.633      & 0.632      \\
$M$                          & 4          & 2          & 3         & 3         & 11         & -          & -          & -         & -          & -          \\
Width $n$                  & 600        & 250        & 350       & 350       & 350        & -          & -          & -         & -          & -          \\
$M_1$                      & -          & -          & -         & -         & -          & 5          & 5          & 2         & 3          & 1          \\
$M_2$                      & -          & -          & -         & -         & -          & 3          & 1          & 1         & 5          & 1          \\
Width $n_1$                & -          & -          & -         & -         & -          & 100        & 100        & 60        & 100        & 60         \\
Width $n_2$                & -          & -          & -         & -         & -          & 60         & 100        & 40        & 80         & 40         \\
$\gamma_1$ ($l_1$ penalty) & $10^{-10}$ & $10^{-20}$ & $10^{-5}$ & $10^{-5}$ & $10^{-5}$  & $10^{-5}$  & $10^{-10}$ & $10^{-5}$ & $10^{-10}$ & $10^{-20}$ \\
$\gamma_2$ ($l_2$ penalty) & $10^-20$   & $10^{-10}$ & $10^{-5}$ & $10^{-5}$ & $10^{-10}$ & $10^{-20}$ & $10^{-20}$ & $10^-20$  & $10^{-3}$  & $10^{-10}$ \\
Dropout rate               & $10^{-3}$  & $10^{-5}$  & $10^{-3}$ & $10^{-3}$ & $10^{-5}$  & $10^{-3}$  & 0.1        & $10^{-3}$ & 0.1        & $10^{-3}$  \\
Batch normalization        & True       & False      & True      & True      & False      & True       & True       & False     & True       & True       \\
Learning rate              & 0.01       & $10^{-3}$  & $10^{-3}$ & $10^{-3}$ & $10^{-4}$  & 0.01       & $10^{-3}$  & 0.01      & $10^{-4}$  & 0.01       \\
Num of iteration           & 10000      & 500        & 5000      & 5000      & 20000      & 20000      & 500        & 5000      & 20000      & 20000      \\
Mini-batch size            & 200        & 500        & 500       & 500       & 100        & 500        & 500        & 200       & 50         & 1000       \\ \hline
\end{tabular}
}
\end{table}

\section*{Appendix IV. Alternative-Specific Connectivity Design and Other Reglarizations in SGP Validation Set}
\label{appendix:regularizations_validation}
\begin{figure}[htb]
\subfloat[$l_1$ Regularization]{\includegraphics[width=0.25\linewidth]{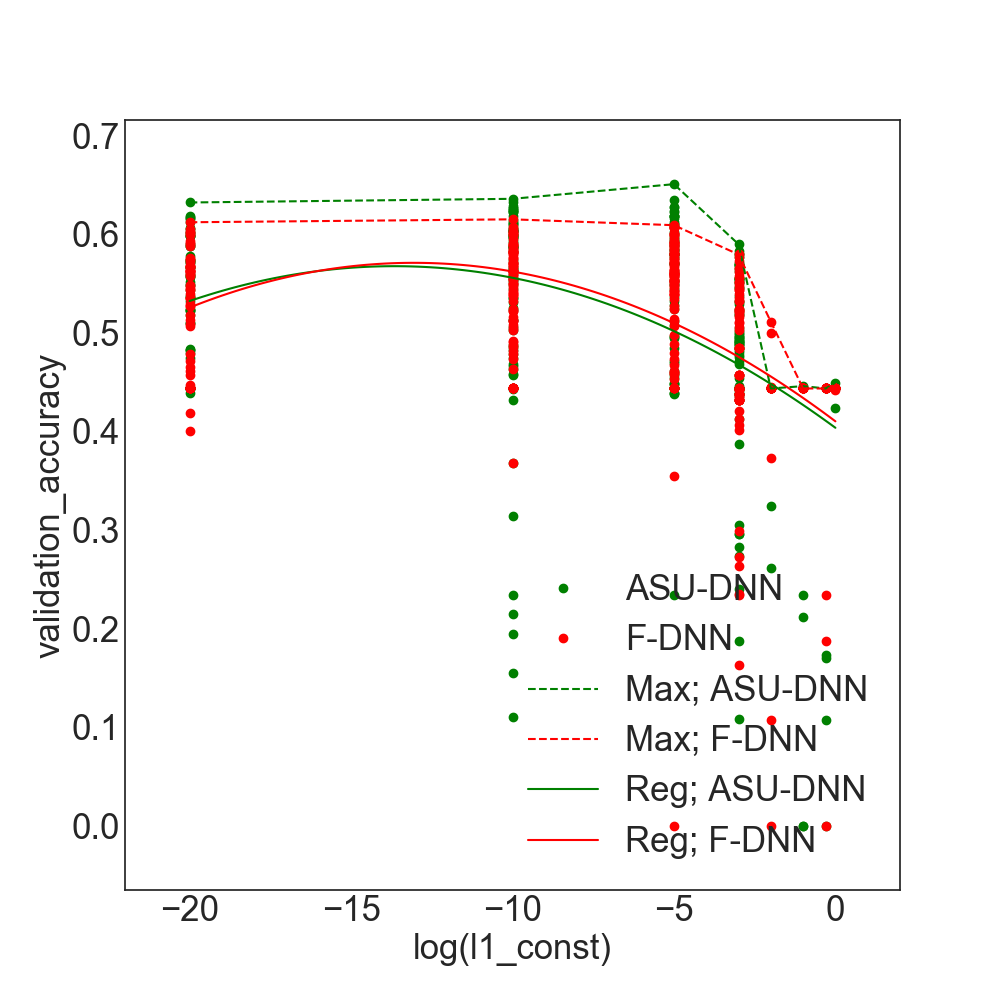}\label{sfig:l1_val}}
\subfloat[$l_2$ Regularization]{\includegraphics[width=0.25\linewidth]{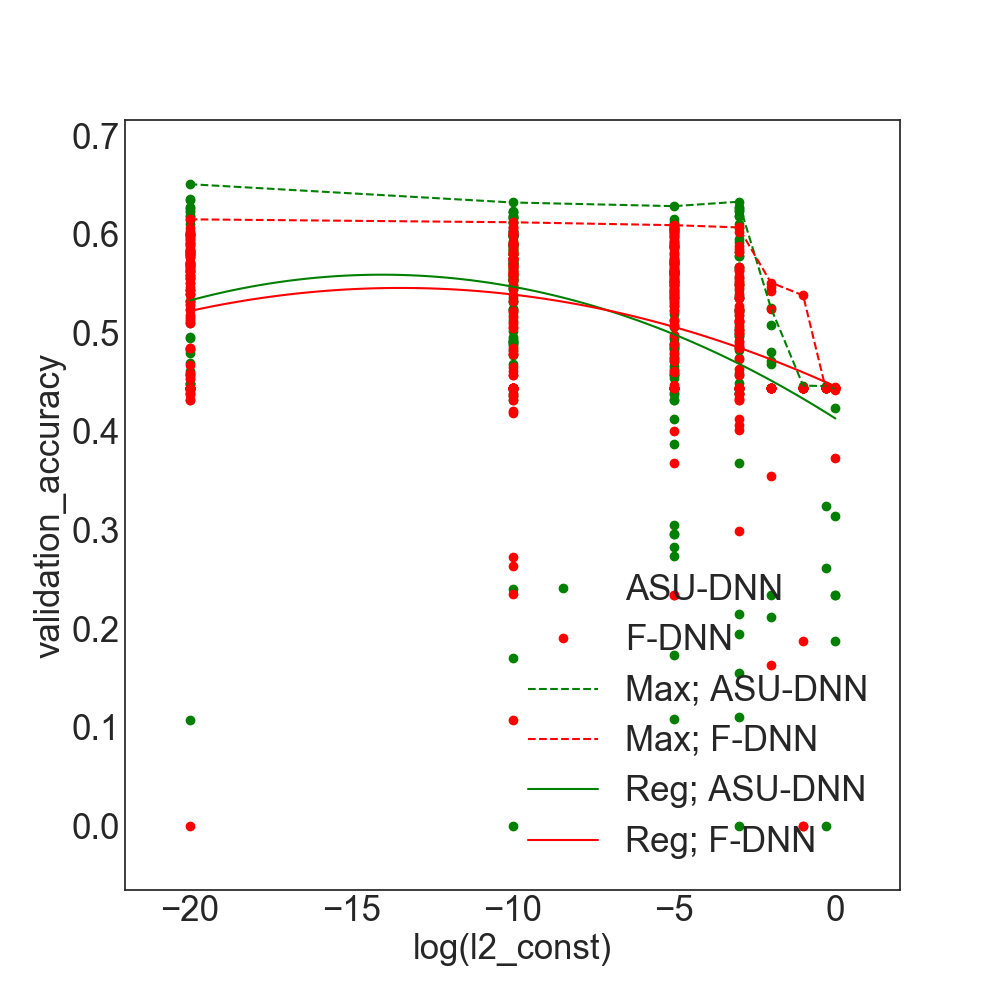}\label{sfig:l2_val}} \\
\subfloat[Learning Rates]{\includegraphics[width=0.25\linewidth]{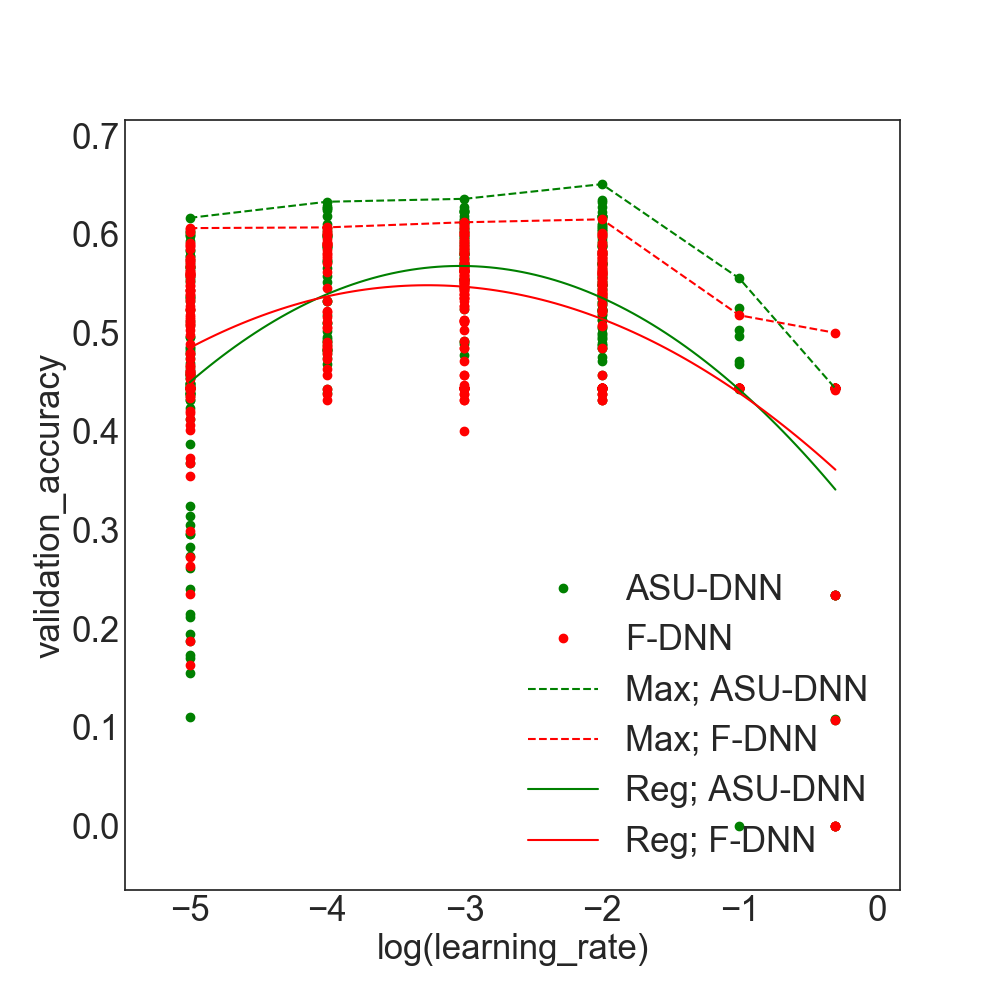}\label{sfig:lr_val}}
\subfloat[Number of Iteration]{\includegraphics[width=0.25\linewidth]{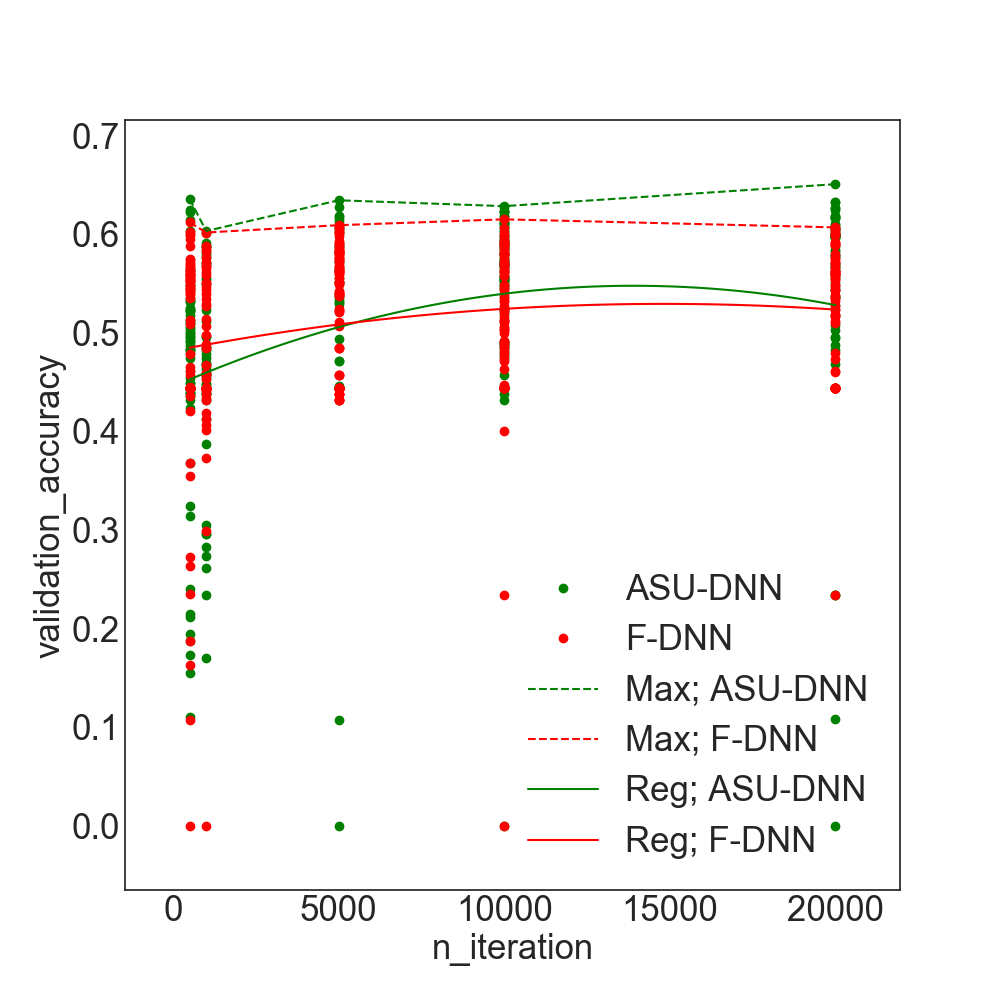}\label{sfig:ni_val}}
\subfloat[Size of Mini Batch]{\includegraphics[width=0.25\linewidth]{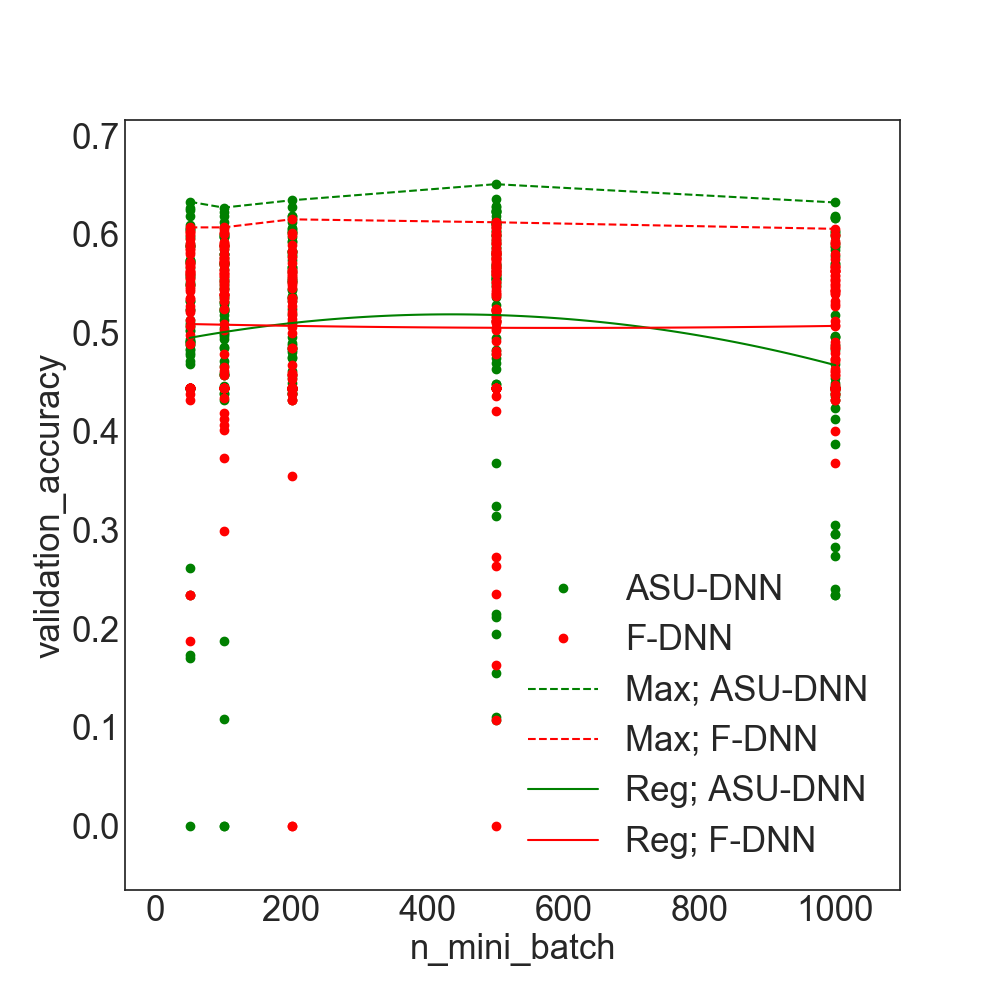}\label{sfig:n_mini_val}}
\subfloat[Batch Normalization]{\includegraphics[width=0.25\linewidth]{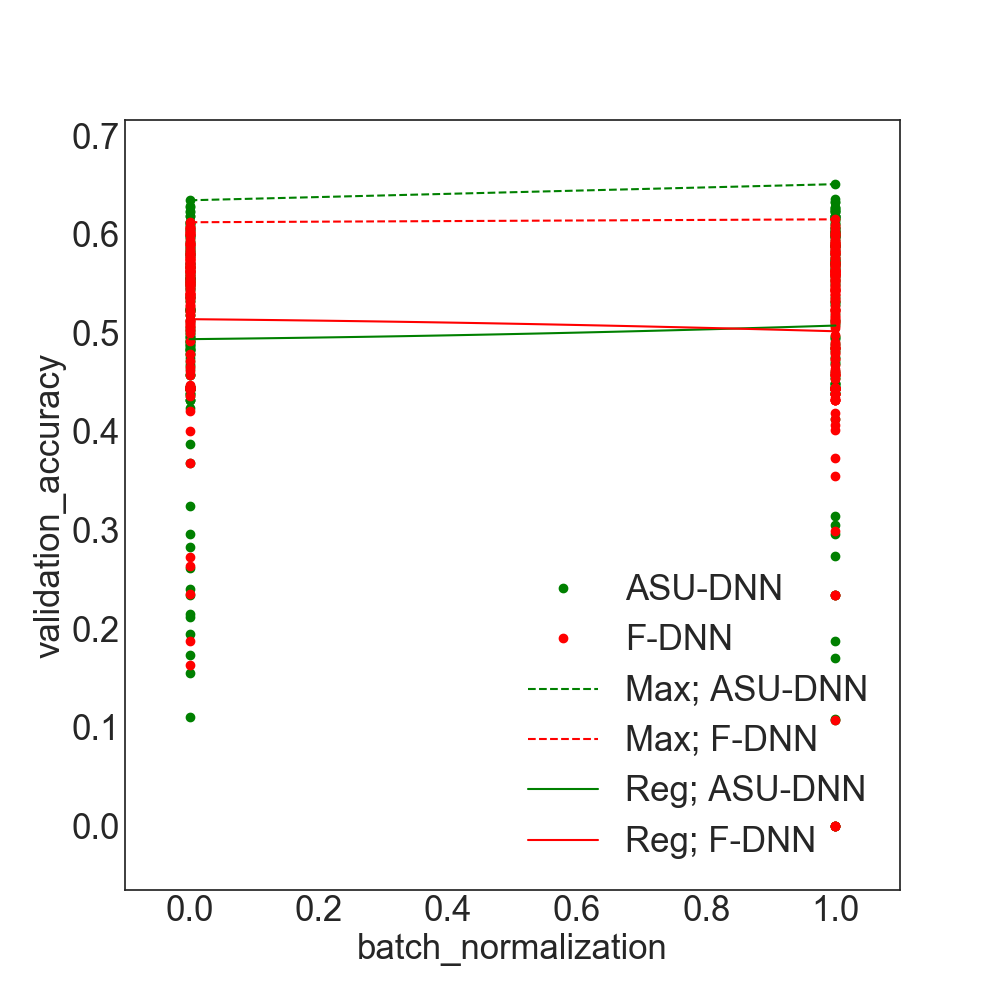}\label{sfig:bn_val}}\\
\subfloat[Depth of DNN]{\includegraphics[width=0.25\linewidth]{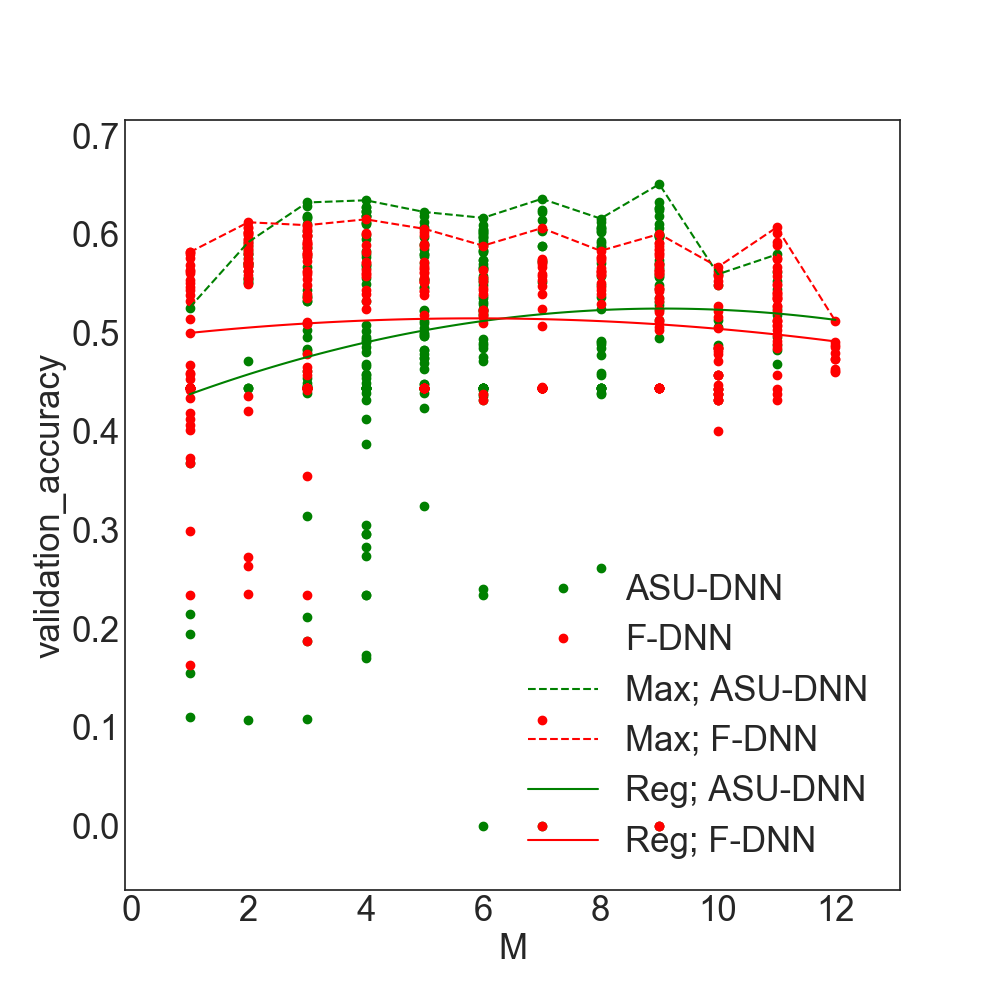}\label{sfig:M_val}}
\subfloat[Width of DNN]{\includegraphics[width=0.25\linewidth]{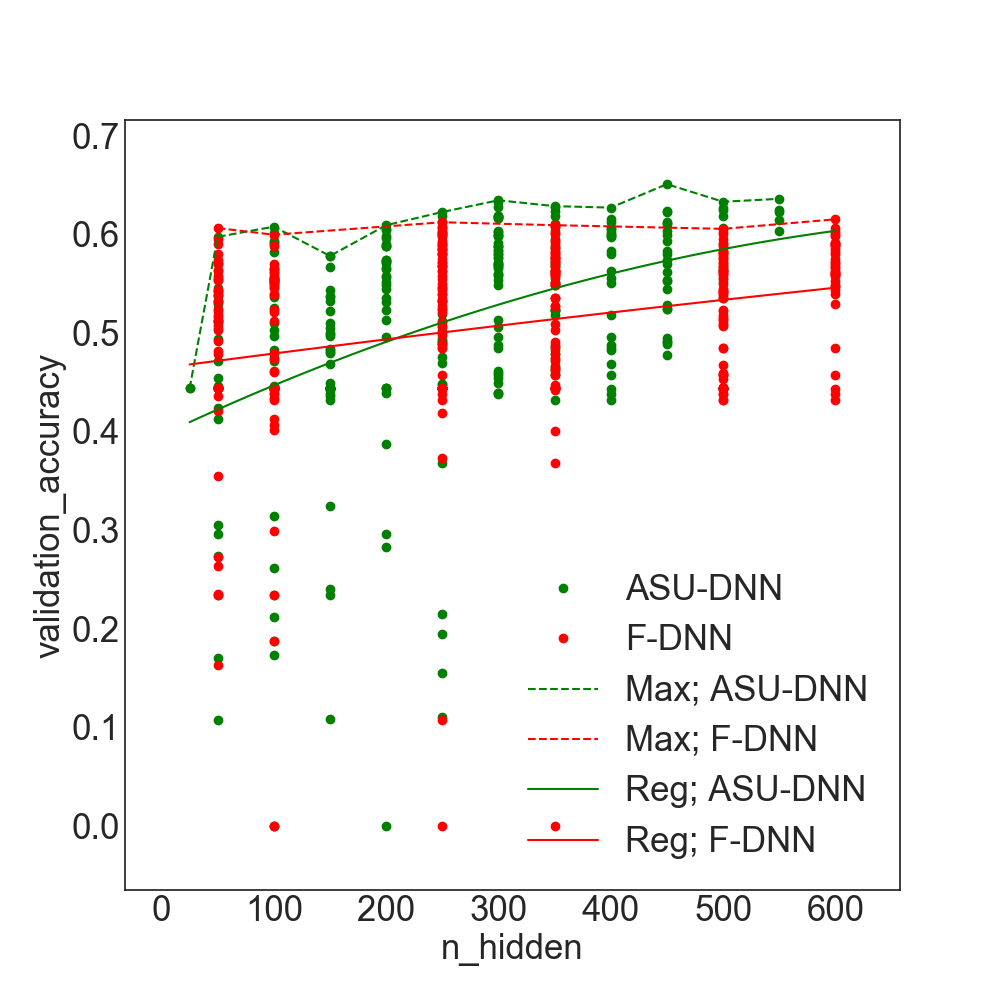}\label{sfig:n_hidden_val}}
\subfloat[Dropout Rates]{\includegraphics[width=0.25\linewidth]{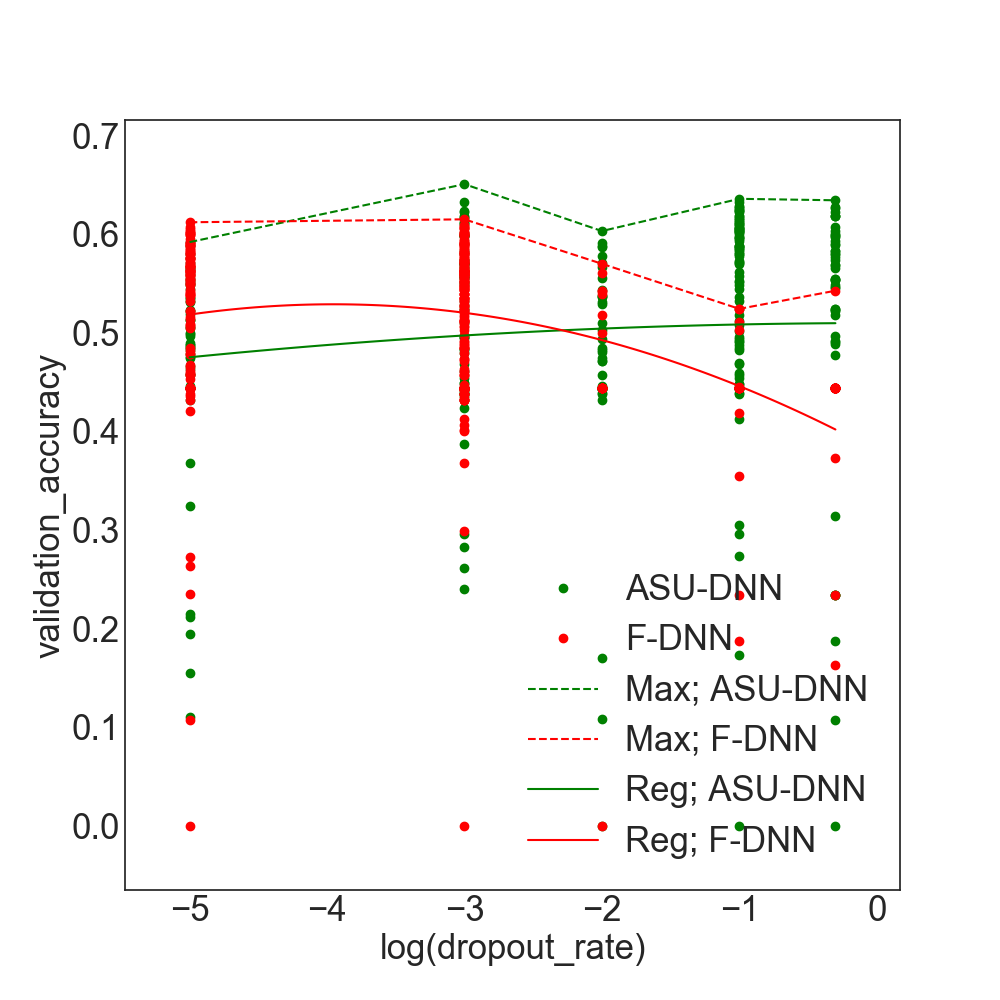}\label{sfig:dr_val}} \\
\caption{Comparing Alternative-Specific Connectivity to Explicit Regularizations, Implicit Regularizations, and Architectural Hyperparameters in SGP Validation Set; \textit{First Row}: Explicit regularizations; \textit{Second Row}: Implicit regularizations; \textit{Third Row}: Architectural hyperparameters. All results are similar to those in testing set.}
\label{fig:ah_appendix}
\end{figure}

\noindent
Figure \ref{fig:ah_appendix} compares the alternative-specific connectivity regularization to other regularization methods in the validation set of SGP. The results are very similar to Figure \ref{fig:ah}.

\end{document}